\documentclass{article}

\usepackage{amsmath,amsfonts,bm}

\def\eqref#1{eq.~\ref{#1}}
\def\Eqref#1{Eq.~\ref{#1}}

\def\1{\bm{1}}

\def\vtheta{{\bm{\theta}}}

\def\vs{{\bm{s}}}

\def\vx{{\bm{x}}}

\def\vz{{\bm{z}}}

\def\mF{{\bm{F}}}

\def\mJ{{\bm{J}}}

\DeclareMathAlphabet{\mathsfit}{\encodingdefault}{\sfdefault}{m}{sl}
\SetMathAlphabet{\mathsfit}{bold}{\encodingdefault}{\sfdefault}{bx}{n}

\usepackage{microtype}
\usepackage{graphicx}
\usepackage{subfigure}
\usepackage{booktabs} 
\usepackage{wrapfig}
\usepackage{siunitx}
\usepackage{multirow}

\usepackage{hyperref}
\usepackage{comment}

\usepackage[accepted]{icml2024}

\usepackage{amsmath}
\usepackage{amssymb}
\usepackage{mathtools}
\usepackage{amsthm}
\newcommand\norm[1]{\left\lVert#1\right\rVert}

\usepackage[capitalize,noabbrev]{cleveref}

\usepackage{bm}
\newcommand{\ourmethodname}{MTF} 
\newcommand{\ournetworkname}{dendPLRNN}

\theoremstyle{plain}

\theoremstyle{definition}

\theoremstyle{remark}

\usepackage[textsize=tiny]{todonotes}

\icmltitlerunning{Integrating Multimodal Data for Joint Generative Modeling of Complex Dynamics}

\begin{document}

\twocolumn[

\icmltitle{Integrating Multimodal Data for Joint Generative Modeling of Complex Dynamics}

\icmlsetsymbol{equal}{*}

\begin{icmlauthorlist}
\icmlauthor{Manuel Brenner}{affil1,affil2}
\icmlauthor{Florian Hess}{affil1,affil2}
\icmlauthor{Georgia Koppe}{equal,affil3,affil5}
\icmlauthor{Daniel Durstewitz}{equal,affil1,affil2,affil3}
\end{icmlauthorlist}

\icmlaffiliation{affil1}{Dept. of Theoretical Neuroscience, Central Institute of Mental Health (CIMH), Medical Faculty Mannheim, Heidelberg University, Germany}
\icmlaffiliation{affil2}{Faculty of Physics and Astronomy, Heidelberg University, Germany}
\icmlaffiliation{affil3}{Interdisciplinary Center for Scientific Computing, Heidelberg University, Germany}
\icmlaffiliation{affil5}{Hector Institute for AI in Psychiatry, CIMH}

\icmlcorrespondingauthor{Manuel Brenner}{manuel.brenner@zi-mannheim.de}
\icmlcorrespondingauthor{Daniel Durstewitz}{daniel.durstewitz@zi-mannheim.de}

\icmlkeywords{Machine Learning, ICML, dynamical systems, chaos, generative models, scientific machine learning, recurrent neural networks, variational autoencoders, exploding gradient problem, teacher forcing}

\vskip 0.3in
]

\printAffiliationsAndNotice{\icmlEqualContribution} 

\begin{abstract}
Many, if not most, systems of interest in science are naturally described as nonlinear dynamical systems. Empirically, we commonly access these systems through time series measurements. Often such time series may consist of discrete random variables rather than continuous measurements, or may be composed of measurements from multiple data modalities observed simultaneously. For instance, in neuroscience we may have behavioral labels in addition to spike counts and continuous physiological recordings. 
While by now there is a burgeoning literature on deep learning for dynamical systems reconstruction (DSR),
multimodal data integration has hardly been considered in this context. Here we provide such an efficient and flexible algorithmic framework that rests on a multimodal variational autoencoder for generating a sparse teacher signal that guides training of a reconstruction model, exploiting recent advances in DSR training techniques. It enables to combine various sources of information for optimal reconstruction, even allows for reconstruction from 
symbolic data (class labels) alone, and connects different types of observations within a common latent dynamics space. In contrast to previous multimodal data integration techniques for scientific applications, our framework is fully \textit{generative}, producing, after training, trajectories with the same geometrical and temporal structure as those of the ground truth system.
\end{abstract}

\section{Introduction}\label{sec:introduction}

For many temporally evolving 
complex systems in physics, biology, or the social sciences, we have only limited knowledge about the generating dynamical mechanisms. Inferring these from data is a core interest in any scientific discipline. It is also practically highly relevant for predicting important changes in system dynamics, like tipping points in climate systems \citep{bury_deep_2021,patel_using_2023}. In recent years, a variety of machine learning (ML) methods for recovering dynamical systems (DS) directly and automatically from time series observations have been proposed \citep{brunton_discovering_2016, vlachas_data-driven_2018, lusch_deep_2018, pathak_model-free_2018, koppe_identifying_2019, schmidt_identifying_2021, fu_dynamically_2019, vlachas2020backpropagation, gauthier_next_2021, jordana_learning_2021, lejarza_data-driven_2022, brenner22a, pmlr-v202-hess23a, chen2023deep, yang2023learning, vlachas2022multiscale}, mostly based on recurrent neural networks (RNNs) for approximating the flow of the underlying true DS \citep{vlachas2020backpropagation, koppe_identifying_2019, schmidt_identifying_2021, brenner22a, pmlr-v202-hess23a, vlachas2022multiscale}. However, almost all of these methods assume that observed time series come as continuous signals with Gaussian noise from a single type of source.  
Yet, time series data from discrete random processes, like class labels or counts, are in fact quite commonplace in many areas, e.g., in the medical domain (electronic health records, smartphone-based data) \citep{koppe_recurrent_2019}, neuroscience (behavioral responses)
\citep{schneider_learnable_2023}, or climate science (event counts) \citep{tziperman-97}. Moreover, with the increasing availability of massive data acquisition techniques in many scientific disciplines, often one has simultaneous measurements from multiple different types of data channels with different 
statistical properties. For instance, in neuroscience one may have simultaneous recordings of spike events and calcium imaging together with behavioral class labels.

\paragraph{Multimodal data integration } 
In general, multimodal data integration is a thriving 
topic in many areas of artificial intelligence and ML research \citep{ahuja2017multimodal, bhagwat_modeling_2018, baltrusaitis_multimodal_2019,  sutter_generalized_2021, shi_relating_2021,  antelmi_multi-channel_2018,liang2022foundations, lipkova_artificial_2022, xu_multimodal_2023,  steyaert2023multimodal, warner2023multimodal, openai2023gpt}. Generative models fusing multiple data channels produce a common latent code which allows for cross-modal prediction or output generation (e.g., images from language; \citet{amrani_noise_2021, radford_learning_2021}), improves inference by complementing information too noisy or missing in one channel through recordings from other modalities \citep{qian2022deep}, or may reveal interesting links among observed modalities \citep{liang_integrative_2015}. 

Variational autoencoders (VAE) \citep{kingma_auto-encoding_2014, rezende_stochastic_2014} are one popular variant of generative models which naturally lend themselves to multimodal settings \citep{baltrusaitis_multimodal_2019, wu_multimodal_2018, sutter_generalized_2021} and a sequential formulation \citep{bayer_mind_2021, girin_dynamical_2021, bai_contrastively_2021}. Longitudinal autoencoders have been proposed \citep{ramchandran_longitudinal_2021} to model temporal correlations in latent space, and have also been extended to multimodal data \citep{ogretir_variational_2022} for the purpose of time series forecasting \citep{antelmi_multi-channel_2018, bhagwat_modeling_2018, dezfouli_integrated_2018, shi_relating_2021, sutter_generalized_2021, qian2022deep}. 

\paragraph{Dynamical Systems Reconstruction (DSR) }
DSR, however, goes beyond mere forecasting in that we request a dynamical model of the data-generating process which captures the temporal and geometrical structure of the true underlying system \citep{koppe_identifying_2019, abarbanel_statistical_2022, platt2023constraining}. A successful DS reconstruction can reveal important invariant topological and geometrical characteristics of the state space the DS lives in, its long-term temporal behavior (attractors), and its sensitivity to perturbations and parameter variations. It provides a mechanistic surrogate for the observed system which can be further analyzed, simulated, perturbed, or lesioned to obtain additional insight into its inner workings \citep{durstewitz_reconstructing_2023}. 

Often special training techniques \citep{mikhaeil_difficulty_2022, brenner22a, pmlr-v202-hess23a, abarbanel2013predicting, abarbanel_statistical_2022, vlachas2023learning},
regularization approaches \citep{schmidt_identifying_2021, platt_systematic_2022, jiang2023training, platt2023constraining}, or structural priors as in physics-informed neural networks (PINNs; \citet{raissi_physics-informed_2019}), are required to achieve these goals. Plain `out-of-the-box' gradient descent techniques or standard variational inference will often fail to reconstruct the underlying system's long-term properties \citep{arribas_rescuing_2020, brenner22a, pmlr-v202-hess23a, platt2023constraining, jiang2023training,durstewitz_reconstructing_2023}, as these standard procedures do not sufficiently constrain the vector field to prevent true and reconstructed trajectories from quickly diverging. This is especially a problem in chaotic systems, where it is related to the exploding or vanishing gradient problem in RNN training \citep{hochreiter_lstm_97, hochreiter2001gradient, bengio_learning_1994, mikhaeil_difficulty_2022, pmlr-v202-hess23a}. Control-theoretic 
techniques are often used to keep model-generated trajectories on track, or to synchronize them with those produced by the observed system, whilst training \citep{voss_nonlinear_2004, abarbanel2009dynamical, abarbanel2013predicting, verzelli2021learn, platt2021robust,abarbanel_statistical_2022}.
Similar ideas underlie the recently proposed methods of sparse \citep{mikhaeil_difficulty_2022, brenner22a} and generalized \citep{pmlr-v202-hess23a} teacher forcing (STF and GTF, respectively) for learning chaotic DS, which balance model-generated and data-inferred states during training in a way that optimally controls gradient flows and trajectory divergence \citep{mikhaeil_difficulty_2022,pmlr-v202-hess23a, eisenmann2023bifurcations}. 

\paragraph{Specific contributions}
Although important in many areas of science, an efficient framework for multimodal data integration \textit{for the purpose of DS reconstruction} is currently essentially lacking. 
A particular challenge here is to reconcile highly efficient training techniques for DS reconstruction like STF or GTF, which usually rest on an invertible decoder model, with the need to infer a common latent code from many different types of data. Here we address this challenge through a novel formulation of the multimodal data integration problem for DS reconstruction: 
Building on the success of VAEs for multimodal integration \citep{baltrusaitis_multimodal_2019, wu_multimodal_2018, sutter_generalized_2021}, we use multimodal VAEs (MVAEs) to construct a common latent representation from random variables following different distributional models. 
Crucially, however, rather than constructing a sequential VAE (SVAE) process that directly operates on this latent code \cite{kramer22a}, 
here we employ the MVAE \textit{to create a multimodal 
TF signal} 
for guiding a DSR model in training. 
Special loss terms and shared decoder models ensure consistency between the MVAE's and reconstruction model's latent codes. This not only 
efficiently exploits the information available in multiple different data streams for DS reconstruction (as compared to a variety of alternative techniques), but for the first time, to our knowledge, enables DS reconstructions of chaotic systems from 
categorical information alone.

\section{Multimodal Teacher Forcing (MTF)}\label{sec:mtf}

\begin{figure*}[!htb]
    \centering
	\includegraphics[width=0.99\linewidth]{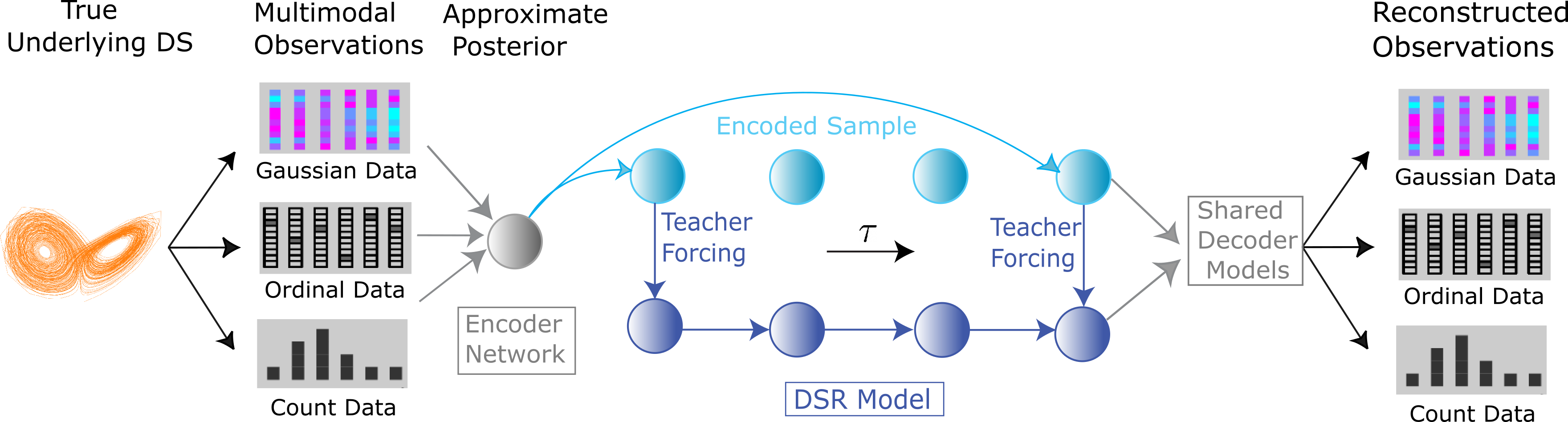}
	\caption{\ourmethodname\ setup. Multimodal observations are translated via an encoder into a common latent representation, which is used for sparse TF in the 
 DSR model's latent space. The latent trajectory is then mapped back into 
 observation space via modality-specific decoder 
 models, which are shared between the MVAE and DSR model.}
	\label{fig:multimodalteacherforcing}
\end{figure*}

Techniques like STF \citep{brenner22a,mikhaeil_difficulty_2022} or GTF \citep{pmlr-v202-hess23a} enable reconstructions even from challenging real-world data on which many previous methods failed. Hence, the key idea for making DS reconstruction from multiple, statistically distinct data sources work, is the generation of a \textit{multimodal TF signal} for efficient training of a reconstruction model. For this we utilize an MVAE, which is then trained jointly with the DSR model through a set of shared decoder models. This general framework, illustrated in Fig. \ref{fig:multimodalteacherforcing}, is very flexible and modular. In principle it can be used with any type of DSR model for approximating the flow of the observed system, as well as with any set of decoder and encoder models. 

\paragraph{DSR model}
While completely flexible, for our specific experiments we chose the recently introduced `dendritic piecewise linear RNN (dendPLRNN)' \citep{brenner22a} as a reconstruction model, as it is both expressive and has a mathematically tractable, piecewise linear design \citep{monfared_transformation_2020, eisenmann2023bifurcations} (other types of RNNs, such as LSTMs or GRUs, may be used and give similar results, see Appx. Table \ref{table_rnns}). The dendPLRNN is defined by the $M$-dimensional latent process equation
\begin{equation}
\label{eq:plrnn_lat}
 \begin{aligned}
     \bm{z}_t =& \bm{A} \bm{z}_{t-1} + \bm{W} \sum_{b=1}^B \alpha_b \max (\bm{0}, \bm{z}_{t-1} - \bm{h}_b)  \\
     &+\bm{h} + \bm{U} \bm{s}_t + \bm{\epsilon}_t, \quad \bm{\epsilon}_t \sim\mathcal{N}(\bm{0},\bm{\Sigma}),
\end{aligned}   
\end{equation}
which describes the temporal evolution of an $M$-dimensional latent state vector $\bm{z}_t=(z_{1t} \dots z_{Mt})^T$ with diagonal matrix $\bm{A} \in \mathbb{R}^{M \times M}$, off-diagonal matrix $\bm{W} \in \mathbb{R}^{M \times M}$, noise covariance matrix $\bm{\Sigma} \in \mathbb{R}^{M \times M}$, 
a nonlinearity given by the `dendritic' spline expansion
with slopes $\alpha_b \in \mathbb{R}$ and thresholds $\bm{h}_b \in \mathbb{R}^M$ \citep{brenner22a}, and a linear term $\bm{U} \in \mathbb{R}^{M \times P}$ weighing $P$-dimensional external inputs $\bm{s}_t$. 

\paragraph{Decoder models}
 To infer this latent process equation jointly from multiple data modalities, the dendPLRNN is connected to different decoder models that take the distinct distributional properties of each modality into account. Our formulation is general, i.e. can work with 
 any combination of continuous and/or discrete data types, and the decoders 
may take the form of any 
differentiable parametric probability model $p_{\theta}(\bullet | \bm{Z})$. For our exposition, however, we consider as concrete example time series of multivariate Gaussian ($\bm{X}$), ordinal ($\bm{O}$),\footnote{Ordinal data, as often encountered in psychology or economics \citep{likert_technique_1932}, are non-metric but still ordered.} and count ($\bm{C}$) nature of length $T$, $\bm{Y}=\{ \{\bm{x}_1,\dots,\bm{x}_T\}; \{\bm{o}_1,\dots,\bm{o}_T\}; \{\bm{c}_1,\dots,\bm{c}_T\} \}$, where modalities $\bm{x}_t \in \mathbb{R}^{N_1}$, $\bm{o}_t \in \mathbb{Z}^{N_2}$, and $\bm{c}_t \in \mathbb{N}^{N_3}$ may all have different dimensions. In this case, we may take as decoders a linear Gaussian model, a cumulative link model for ordinal data, and a log-link function for Poisson data:
\begin{align}
 \bm{x}_t | \bm{z}_t &\sim \mathcal{N}\left(\bm{B} \bm{z}_t, \bm{\Gamma}\right); \label{eq:gauss_obs_model} \\ 
 \bm{o}_t | \bm{z}_t &\sim \operatorname{Ordinal}(\bm{\beta} \bm{z}_t, \epsilon ); \label{eq:ordinal_obs_model} \\ 
 \bm{c}_t | \bm{z}_t &\sim \operatorname{Poisson}(\lambda(\bm{z}_t)), \label{eq:poisson_obs_model}
\end{align}
The DSR model loss is now given by the sum of the negative log-likelihoods of the decoders, assuming conditional independence given latent states $\bm{z}$:
\begin{equation}\label{loss_DSR_mtf}
\begin{aligned}
\mathcal{L}_\text{DSR}=-\sum_{t=1}^{T}(\log p_{\bm{\theta}}(\bm{{x}}_t|\bm{z}_{1:K,t})+ \\ \log p_{\bm{\theta}}(\bm{o}_t|\bm{z}_{1:K,t})+\log p_{\bm{\theta}}(\bm{c}_t|\bm{z}_{1:K,t})),
\end{aligned}
\end{equation}
where only the first $K$ latent states are used to allow for sharing with the MVAE as explained below, and the individual likelihood terms are specified in Appx. \ref{sec:methods:obsmodels}.

\paragraph{Training: Multimodal Teacher Forcing (MTF) }
Training RNNs on time series is generally challenging due to the exploding and vanishing gradient problem \citep{bengio_learning_1994}.
This becomes particularly severe in DS reconstruction as for chaotic systems exploding gradients cannot be avoided even in principle, due to the positive maximum Lyapunov exponent resulting in exponential divergence of trajectories \citep{mikhaeil_difficulty_2022,pmlr-v202-hess23a}. Yet, at the same time batch length cannot be too short, as the training algorithm will then fail to capture the system's long term behavior and limit sets \citep{brenner22a, platt_systematic_2022, platt2023constraining}. STF \citep{mikhaeil_difficulty_2022} and GTF \citep{pmlr-v202-hess23a} therefore balance loss and trajectory divergence with the need to capture relevant long time scales in an optimal way. They do so by replacing forwarded-iterated latent states $\bm{z}_t$ by data-inferred states $\bm{\hat z}_t$ sparsely, that is only at strategically chosen time points determined from the empirically estimated maximum Lyapunov exponent in the case of STF \citep{mikhaeil_difficulty_2022}, or by averaging forward-propagated and data-inferred states in an optimal way in the case of GTF \citep{pmlr-v202-hess23a}. Data-inferred states are obtained by inverting the decoder model (e.g., $\bm{\hat z}_t = (\bm{B}^T \bm{B})^{-1} \bm{B}^T \bm{x}_t$ for a simple linear-Gaussian model as in \Eqref{eq:gauss_obs_model}, i.e. taking the pseudo-inverse). This forcing is then only applied during model training, not at run time (see Methods \ref{sec:method:mtf} for further details).

Inverting the decoder model is in general not possible, however, in particular for discrete random variables. Moreover, in the case of multiple simultaneously observed data modalities, it is unclear how to combine the different data modalities to obtain an optimal estimate $\bm{\hat z}_t$. We therefore need another means to create a (sparse) TF signal if we would like to utilize STF or GTF for training. Here we achieve this through an MVAE for building a joint latent representation over the different data types, $\bm{\tilde{z}}_t \sim p(\bm{\tilde{z}}_t | \bm{X}, \bm{O}, \bm{C})$, which can be used as a TF signal at time $t$. We denote the MVAE states by $\bm{\tilde{z}}_t\in \mathbb{R}^{K}$ to avoid confusion with the latent dynamical process $\bm{z}_t\in \mathbb{R}^{M}$ generated by the reconstruction model.\footnote{Note that in general we do not assume the MVAE and reconstruction model latent codes to have the same dimensionality, i.e. $K \leq M$, 
which, e.g., could enable the DSR model to capture additional, unobserved latent states or non-stationarity in the data; see Methods \ref{sec:method:mtf} for further details.} 
To ensure consistency between the DSR model's and the MVAE's latent states, the MVAE is coupled to the observations through the \textit{very same} set of decoder models
\begin{align} 
\bm{x}_t | \bm{\tilde{z}}_t &\sim \mathcal{N}\left(\bm{B} \bm{\tilde{z}}_t, \bm{\Gamma}\right); \label{eq:mvae_gauss_obs_model} \\
\bm{o}_t | \bm{\tilde{z}}_t &\sim \operatorname{Ordinal}(\bm{\beta} \bm{\tilde{z}}_t, \epsilon );
\label{eq:mvae_ordinal_obs_model} \\
\bm{c}_t | \bm{\tilde{z}}_t &\sim \operatorname{Poisson}(\lambda(\bm{\tilde{z}}_t)) \label{eq:mvae_poisson_obs_model},  
\end{align}
\textit{sharing} all decoder parameters (i.e. $\bm{B}, \bm{\Gamma}, \bm{\beta}$ etc.) with the DSR model (see Fig. \ref{fig:multimodalteacherforcing}). 

The MVAE is trained by minimizing the negative Evidence Lower Bound (ELBO)
\begin{equation}
\label{eq:elbo}
\begin{aligned} 
\mathcal{L}(\bm{\phi},\bm{\theta}; \bm{Y})= &-\mathbb{E}_{q_{\bm{\phi}}}[\log p_{\bm{\theta}}(\bm{Y}|\bm{\tilde{Z}}) \\ & +\log p_{\bm{\theta}}(\bm{\tilde{Z}})] 
-\mathbb{H}_{q_{\bm{\phi}}}(\bm{\tilde{Z}} | \bm{Y})&
\end{aligned}
\end{equation}

using the reparameterization trick for latent random variables \citep{kingma_auto-encoding_2014}, where $\mathbb{H}_{q_{\bm{\phi}}}$ is the entropy of the approximate posterior, and assuming conditional independence of the observations given the latent states. We tested various choices for the encoder model ${q_{\bm{\phi}}}(\bm{\tilde{Z}}|\bm{Y})$, including mixture-of-experts, 
RNNs or transformers, but achieved best results with a temporal CNN with all data modalities concatenated (see Table \ref{table_encoders} for results and Methods \ref{sec:method:mtf} for architectural details). 

\paragraph{Process prior and total loss}
We thus have specified ${p_{\bm{\phi}}}(\bm{Y}|\bm{\tilde{Z}})$ through our choice of decoder models, and 
parameterized ${q_{\bm{\phi}}}(\bm{\tilde{Z}}|\bm{Y})$ by a temporal CNN. But what is the best choice for the prior $p_{\bm{\theta}}(\bm{\tilde{Z}})$? Crucially, to further enforce consistency between the MVAE and DSR model latent codes, $\bm{\tilde{Z}}$ and $\bm{Z}$, respectively, we assume that $p_{\bm{\theta}}(\bm{\tilde{Z}})$ is given through the DSR model (the dendPLRNN in our case). Specifically, in line with the Gaussian assumptions in the dendPLRNN, \Eqref{eq:plrnn_lat}, we assume for the second term in \Eqref{eq:elbo}: 
\begin{equation}
\label{eq:prior_term}
\begin{aligned} 
-\mathbb{E}_{q_{\bm{\phi}}}[\log p_{\bm{\theta}}(\bm{\tilde{Z}})] &\approx \frac{1}{L} \sum_{l=1}^L \sum_{t=1}^T \frac{1}{2} \bigl( \log|\bm{\Sigma}| \\ 
 &+(\bm{\tilde{z}}^{(l)}_{t}-\bm{\mu}_{t})^{\top} 
\bm{\Sigma}^{-1} (\bm{\tilde{z}}^{(l)}_{t}-\bm{\mu}_{t})\\
&+const.\bigl),
\end{aligned}
\end{equation}
where the expectation value is approximated by $L$ Monte Carlo samples $\bm{\tilde{z}}^{(l)}_{t} \sim q_{\bm{\phi}}(\bm{\tilde{z}}_t|\bm{Y})$, and the means $\bm{\mu}_{t} = \mathbb{E}(\bm{z}_t|\bm{z}_{t-1})$ directly come from the DSR model (see Methods \ref{sec:method:mtf} for details). We therefore also call this term, \Eqref{eq:prior_term}, the \textit{consistency loss}, $\mathcal{L}_{\text{con}}$. 

The total training loss is now given by the MVAE's negative ELBO, \Eqref{eq:elbo}, combined with the reconstruction loss (negative log-likelihood) of the DSR model (funneled through the same set of decoder models):
\begin{equation}\label{eq:loss_total}
\begin{aligned}
   \mathcal{L}_\text{MTF} = 
   &-\mathbb{E}_{q_{\bm{\phi}}}[\log p_{\bm{\theta}}(\bm{Y}|\bm{\tilde{Z}})] -\mathbb{H}_{q_{\bm{\phi}}}(\bm{\tilde{Z}} | \bm{Y}) \\ &\underbrace{-\mathbb{E}_{q_{\bm{\phi}}}[\log p_{\bm{\theta}}(\bm{\tilde{Z}})]}_{\mathcal{L}_\text{con}} \ \ \underbrace{-\log p_{\bm{\theta}}(\bm{Y}|\bm{Z})}_{\mathcal{L}_\text{DSR}} 
\end{aligned}
\end{equation}

Whilst training, we then use $\bm{\tilde{z}}_t$ as our TF signal for guiding the DSR model. Specifically, for STF we replace (a subset of the) latent states $\bm{z}_t$ by $\bm{\tilde{z}}_t$ at strategically chosen times $l\tau+1,\ l \in \mathbb{N}_0$, as illustrated in Fig. \ref{fig:multimodalteacherforcing} (similarly, for GTF we may take a weighted average; see \citet{pmlr-v202-hess23a} for details).

\begin{table*}
\caption{Comparison of dendPLRNN trained by \ourmethodname\ (proposed method), by a sequential multimodal VAE (SVAE) based on \cite{kramer22a}, a VAE-TF approach similar to \ourmethodname\ except that all data modalities were `Gaussianized' (GVAE-TF), BPTT-TF as in \cite{brenner22a} using Gaussianized data, and a multiple-shooting (MS) approach. 
Training was performed on multivariate normal, ordinal, and count data produced by the chaotic Lorenz system, Rössler system, and Lewis-Glass model. 
Values are mean $\pm$ SEM, averaged over 15 trained models. 
X = value cannot be computed for this model (e.g., because resp. decoder model is not present). Note that SCC (Spearman cross-correlation), OACF (ordinal autocorrelation function), and CACF (count autocorrelation function) all refer to mean-squared-errors (MSEs) between ground truth and generated correlation functions. Bold numbers indicate top performance within $\pm 1$ SEM.\\}
\centering
\renewcommand{\arraystretch}{1.1}
\large
\scalebox{0.5}{
\begin{tabular}{lcccccccc}
\hline
Dataset& Method  & $D_{stsp}$ $\downarrow$ & $D_{H}$ $\downarrow$ & PE $\downarrow$ & OPE $ \downarrow$ & SCC $\downarrow$ & OACF $\downarrow$  & CACF $\downarrow$ \\ \hline
\multirow{6}{*}{Lorenz}  & \ourmethodname\ & $\mathbf{3.4 \pm 0.35}$   & $\mathbf{0.30 \pm 0.06}$  & $\mathbf{1.3\mbox{e$-$}2} \pm \mathbf{2\mbox{e$-$}4}$  & $\mathbf{0.12 \pm 0.03}$  &  $\mathbf{0.07 \pm 0.01}$  &  $\mathbf{0.07 \pm 0.01}$   &  $\mathbf{6.6\mbox{e$-$}5 \pm 8.1\mbox{e$-$}6}$  \\ 
                        & SVAE    & $11.1 \pm 0.6$   & $0.82 \pm 0.05$  & ${6.3\mbox{e$-$}1}$ $\pm$ ${5.1\mbox{e$-$}2}$  & $0.68 \pm 0.03$  &  $0.14 \pm 0.01$ &  $0.18 \pm 0.02$   &  ${8.5\mbox{e$-$}5}$  $\pm$ ${1.6\mbox{e$-$}5}$      \\
                        & BPTT    & $6.31 \pm 1.2$   & $0.47 \pm 0.11$  & ${2.1\mbox{e$-$}1}$ $\pm$ ${2.4\mbox{e$-$}2}$  & $0.33 \pm 0.04$  &  $0.11 \pm 0.02$ &  $0.09 \pm 0.02$   &  ${8.2\mbox{e$-$}5}$ $\pm$ ${9\mbox{e$-$}6}$      \\
                        
                          & MS    & $4.5 \pm 1.5$   & $0.61 \pm 0.08$  & X  & X &  $0.14 \pm 0.04$   & $0.11 \pm 0.02$   & $\mathbf{{6.5\mbox{e$-$}5} \pm {3.8\mbox{e$-$}6}}$ \\
                        
                        & GVAE-TF    & $4.3 \pm 0.3$   & $0.47 \pm 0.07$  & ${3.6\mbox{e$-$}1} \pm {1.5\mbox{e$-$}3}$  & X &  X   & X   &  X      \\
                         & BPTT-TF    & $8.8 \pm 1.9$   & $0.86 \pm 0.05$  & ${4.4\mbox{e$-$}1} \pm {2.2\mbox{e$-$}2}$  & X &  X   & X   &  X      \\  \hline
                 
\multirow{6}{*}{Rössler} &\ourmethodname\ & $\mathbf{1.45 \pm 0.71}$   & $\mathbf{0.32 \pm 0.03}$  & $\mathbf{{1.9\mbox{e$-$}3} \pm {7.1\mbox{e$-$}5}}$  & $\mathbf{0.08 \pm 0.02}$  &  $\mathbf{0.04 \pm 0.004}$  &  $\mathbf{0.017 \pm 0.003}$   &  $\mathbf{{6.5\mbox{e$-$}5} \pm {1.2\mbox{e$-$}5}}$ \\ 
                         & SVAE    & $10.7 \pm 1.5$   & $0.66 \pm 0.05$  & ${1.5\mbox{e$-$}1}$ $\pm$ ${3.1\mbox{e$-$}2}$  & $0.24 \pm 0.02$  &  $0.17 \pm 0.03$ &  $0.13 \pm 0.02$   & ${1.1\mbox{e$-$}4} \pm {1.4\mbox{e$-$}5}$  \\
                          & BPTT    & $9.05 \pm 0.5$   & $0.7 \pm 0.01$  & ${7.4\mbox{e$-$}2}$ $\pm$ ${2.0\mbox{e$-$}3}$  & $0.18 \pm 0.02$  &  $0.3 \pm 0.03$ &  $0.19 \pm 0.07$   &  ${1.5\mbox{e$-$}4}$ $\pm$ ${6\mbox{e$-$}6}$      \\
                         & MS    & $3.99 \pm 1.1$   & $0.59 \pm 0.04$  & X  & X &  $0.08 \pm 0.04$   & $0.09 \pm 0.02$   & ${1.6\mbox{e$-$}4}$ $\pm$ ${5.9\mbox{e$-$}5}$ \\
                         
                         & GVAE-TF    & $12.1 \pm 0.5$   & $0.55 \pm 0.04$  & ${4.9\mbox{e$-$}2} \pm {3.4\mbox{e$-$}3}$  &X  & X  & X &  X  \\
                         & BPTT-TF    & $8.9 \pm 1.4$   & $0.64 \pm 0.07$  & ${2.8\mbox{e$-$}1} \pm {1.8\mbox{e$-$}3}$  & X &  X   & X   &  X      \\ \hline
\multirow{6}{*}{Lewis-Glass} &\ourmethodname\ & $\mathbf{0.27 \pm 0.07}$   & $\mathbf{0.33 \pm 0.02}$  & $\mathbf{{2.1\mbox{e$-$}3} \pm {7\mbox{e$-$}5}}$  & $\mathbf{0.11 \pm 0.01}$  &  ${0.12 \pm 0.03}$  &  $\mathbf{0.05 \pm 0.02}$   &  ${2.3\mbox{e$-$}4} \pm {2.0\mbox{e$-$}5}$ \\ 
                         & SVAE    & $2.6 \pm 0.5$   & $0.52 \pm 0.03$  & ${8.0\mbox{e$-$}2}$ $\pm$ ${4\mbox{e$-$}3}$  & $0.26 \pm 0.01$  &  $0.4 \pm 0.05$ &  $0.18 \pm 0.03$   & ${7.5\mbox{e$-$}3} \pm {4.7\mbox{e$-$}3}$  \\
                        & BPTT    & $2.8 \pm 0.5$   & $0.57 \pm 0.05$  & ${6.2\mbox{e$-$}2}$ $\pm$ ${3\mbox{e$-$}3}$  & $0.23 \pm 0.02$  &  $0.47 \pm 0.08$ &  $0.21 \pm 0.03$   & ${9.1\mbox{e$-$}3} \pm {3.2\mbox{e$-$}3}$  \\
                         
                          & MS    & $\mathbf{0.33 \pm 0.06}$   & $\mathbf{0.35 \pm 0.01}$  & X  & X &  $\mathbf{0.08 \pm 0.01}$   & $\mathbf{0.04 \pm 0.002}$   & $\mathbf{{1.9\mbox{e$-$}4}\pm {7.5\mbox{e$-$}6}}$ \\ 
                         & GVAE-TF    & $\mathbf{0.28 \pm 0.08}$   & $0.44 \pm 0.02$  & ${4.6\mbox{e$-$}3} \pm {4\mbox{e$-$}4}$  &X  & X  & X &  X  \\
                          & BPTT-TF    & $2.51 \pm 0.71$   & $0.43 \pm 0.03$  & ${2.6\mbox{e$-$}2} \pm {3\mbox{e$-$}3}$  & X &  X   & X   &  X      \\ 
\hline
\end{tabular}
}
\label{table_noisy}
\end{table*}

Note that our usage of an MVAE here thus profoundly differs from the more conventional use in, for instance, SVAEs \citep{sutter_generalized_2021, kramer22a}: The MVAE's major purpose is to generate an appropriate TF control signal which \textit{guides} the DSR model in training, not in modeling the process itself as in an SVAE; see Fig. \ref{svae_mtf_comparison} for an illustration of this important difference. 
This control of model-generated trajectories through data-inferred values has been found crucial for successful DSR \cite{mikhaeil_difficulty_2022,brenner22a,pmlr-v202-hess23a}. This in turn necessitates the reconstruction loss $\mathcal{L}_\text{DSR}$, which allows for the DSR model to also make direct contact with the data, rather than only indirectly through the posterior approximation. Without it, much of the burden for recovering the dynamics would be on the (fallible) MVAE, while with it, through the sharing of decoder models, the approximate posterior will be improved through the DSR process. Ablation studies indeed further confirm that all components of the loss (Fig. \ref{fig:loss_scaling}), including $\mathcal{L}_\text{DSR}$, as well as an optimal choice for the TF interval (Fig. \ref{fig:influence_tau}), are crucial to the success of the approach. Specifically, optimal TF intervals were associated with smooth and well navigable loss landscapes (Fig. \ref{fig:losslandscape}).

\section{Results}\label{sec:experiments}
\subsection{Evaluating MTF: Comparison to Other Algorithmic Strategies}
\label{sec:exp:benchmarks}

\begin{figure}[!htb]
    \centering
	\includegraphics[width=0.99\linewidth]{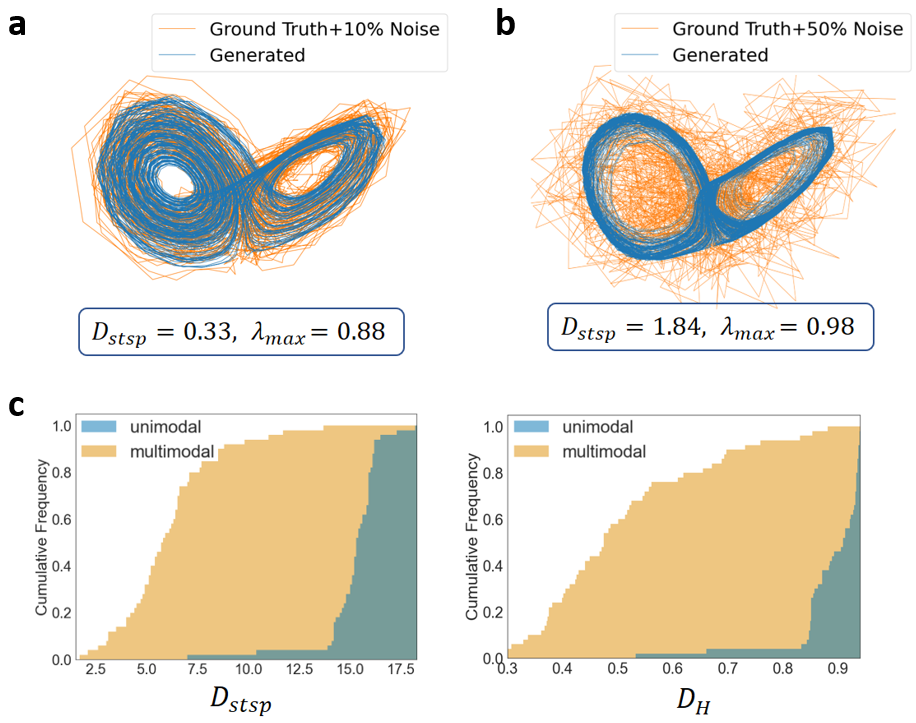}
	\caption{DS reconstruction from moderately (\textbf{a}) and heavily (\textbf{b}-\textbf{c}) distorted continuous observations (Gaussian observation noise of $10 \%$ and $50 \%$, respectively, of the data variance) and other simultaneously provided observation modalities, sampled from a Lorenz-63 system. 
    \textbf{a}: Freely generated example trajectories from a dendPLRNN ($M=20, B=10, K=20, \tau=10$) 
    trained with \ourmethodname\ jointly on Gaussian ($10 \%$ noise), ordinal, and count data. 
    \textbf{b}: Same as \textbf{a} for a dendPLRNN trained by \ourmethodname\
    on heavily distorted Gaussian 
    ($50 \%$ noise) and ordinal observations. Note that even in this case the butterfly wing structure of the Lorenz attractor was successfully captured. 
    The maximum Lyapunov exponent ($\lambda_{\text{max}}$) furthermore confirms the dendPLRNN-generated attractor is chaotic (for the GT Lorenz system, $\lambda_{\text{max}} \approx 0.903$). \textbf{c}: Normalized cumulative histograms of geometrical attractor disagreement ($D_{stsp}$, left) and Hellinger distance ($D_{H}$, right) between reconstructed and ground-truth system for the same setting as in \textbf{b}.}
    \label{fig:noisy_lorenz}
\end{figure}

We first compared MTF on moderately challenging multimodal DS reconstruction tasks\footnote{Moderately challenging because many of the comparison methods actually broke down on the more difficult setups.} to a variety of other possible algorithmic strategies that may be considered, all of which lack one or more key ingredients of our methodology. For this, a training and a test set of $100,000$ time steps were created from a Lorenz-63 and a Rössler system with $1\%$ process noise, and a $6$d Lewis-Glass network model \citep{lewis_nonlinear_1992, gilpin_chaos_2022}, all in their chaotic regimes (see Methods \ref{sec:methods:datasets} for detailed parameter settings and numerical integration). From the simulated trajectories, we then sampled ordinal and count observations using Eqs. \ref{eq:ord_obs} and \ref{eq:poisson_obs} (with randomly drawn parameters), as well as continuous observations with $10\%$ Gaussian noise. Example reconstructions by 
\ourmethodname\ are in Fig. \ref{fig:noisy_lorenz}\textbf{a} and Fig. \ref{fig:example_trajectories}. 

We compared the performance of MTF on these simulated data to the following setups: First, to a multimodal SVAE (\citet{kramer22a}; to our knowledge the only other general approach intended for DSR from multimodal data). 
Second, we tried classical RNN training, where observations are provided as inputs at every time step, via truncated BPTT using modality-specific decoder models, Eqs. \ref{eq:gauss_obs_model}-\ref{eq:poisson_obs_model}. A third approach that can deal with multimodal observation models but, unlike TF, does not require model inversion, is `multiple shooting (MS)', a method suggested in the dynamical systems literature \citep{voss_nonlinear_2004}, but to our knowledge not extended to multimodal data so far. 
We also included two other comparisons where we first transformed multimodal data to be approximately Gaussian (via Box-Cox \& Gaussian kernel smoothing, see Methods \ref{sec:methods:gaussian}), and then either trained the RNN via standard BPTT-TF \citep{brenner22a} or by VAE-TF, similar as in MTF, but without modality-specific decoder models (labeled GVAE-TF). For comparability the same RNN architecture (the dendPLRNN, \Eqref{eq:plrnn_lat}) was used in all these comparisons.

Since in DS reconstruction, we are primarily interested in invariant and long-term properties of the underlying DS \citep{mikhaeil_difficulty_2022, platt2023constraining, jiang2023training, pmlr-v202-hess23a, platt2021robust}, for evaluation we focused on measures that capture the geometrical structure in state space ($D_{\text{stsp}}$, a type of Kullback-Leibler divergence; \citet{koppe_identifying_2019,brenner22a,pmlr-v202-hess23a}) and the asymptotic temporal structure (evaluated either through the Hellinger distance, $D_{\text{H}}$, on power spectra \cite{mikhaeil_difficulty_2022,pmlr-v202-hess23a}, or auto-covariance functions for the ordinal and categorical data; Methods \ref{sec:exp:measures}). We also report mean-squared errors (MSE) to assess short-term ahead prediction ($10$-step-ahead $L_2$ prediction error (PE) for continuous and $L_1$ PE for ordinal data \citep{ogretir_variational_2022}). However, especially in chaotic systems, PEs are not suited for evaluating the system's longer-term behavior because of the exponential divergence of nearby trajectories \citep{wood_statistical_2010,mikhaeil_difficulty_2022}. Details of all measures are provided in Methods \ref{sec:exp:measures}. As evidenced in Table \ref{table_noisy}, \ourmethodname\ outperforms all other possible model setups, and in particular the multimodal SVAE by large margins, reinforcing our point about the difficulty of learning an appropriate posterior in the absence of control-theoretic guidance of the training process (cf. Fig. \ref{svae_mtf_comparison}).

To investigate multimodal data integration in a more challenging setting, we next tested a scenario where continuous observations from the Lorenz-63 system were heavily distorted by Gaussian noise with $50\%$ of the data variance. 
At the same time, ordinal observations 
with $8$ variables divided into $7$ ordered categories each, $o_{nt} \in \{1 \dots 7\}$, $n=1\dots8$, were sampled using \Eqref{eq:ord_obs}. 
We then trained the dendPLRNN via \ourmethodname\ once with, and once without, ordinal observations. Fig. \ref{fig:noisy_lorenz}\textbf{b} proves that with ordinal observations on board, DS reconstruction is, in principle, possible even under these challenging conditions. The cumulative histograms of the geometric measure $D_{stsp}$ 
and the temporal measure $D_{H}$ in Fig. \ref{fig:noisy_lorenz}\textbf{c} furthermore show that inclusion of ordinal observations significantly improves reconstructions. 

\begin{figure}[!htb]
    \centering
	\includegraphics[width=0.99\linewidth]{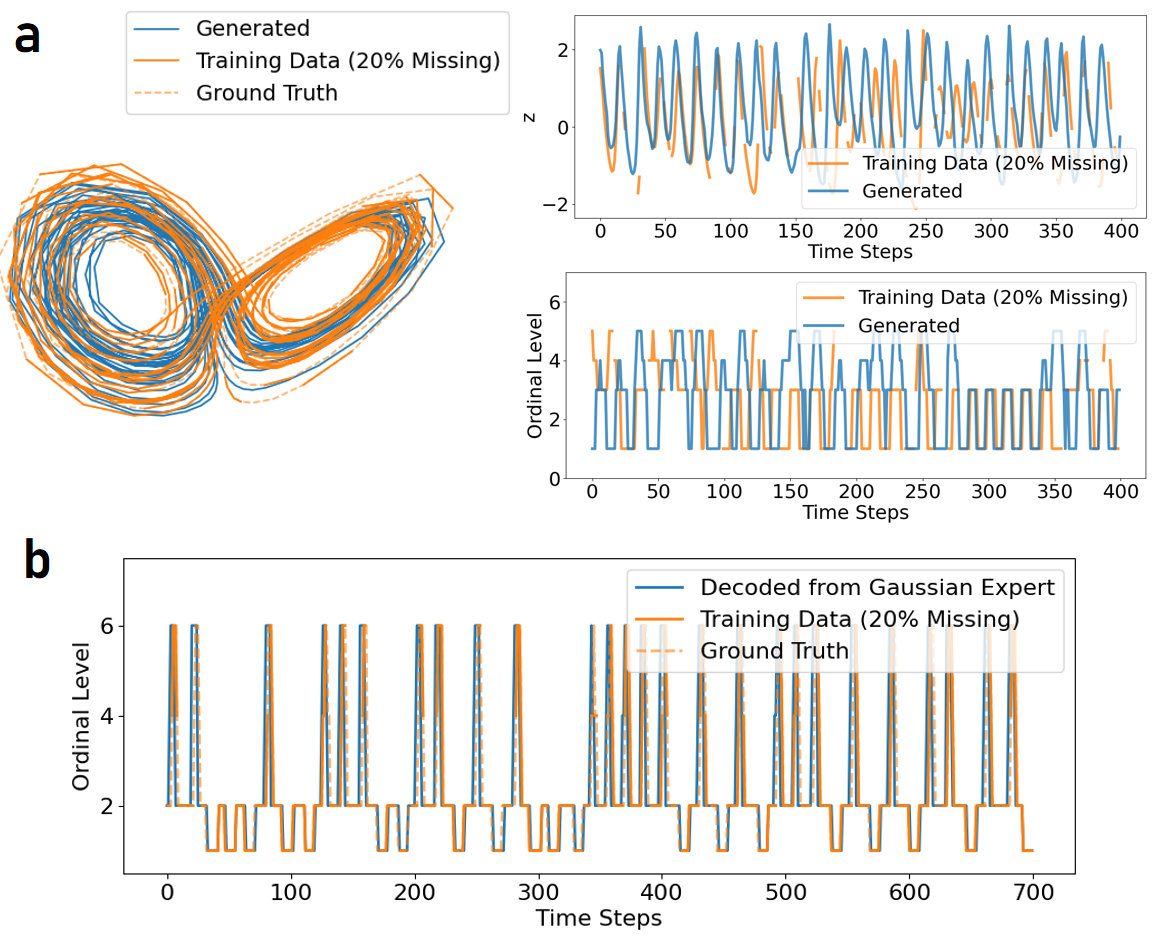}
	\caption{Cross-modal inference with missing observations, using a mixture-of-experts encoder. \textbf{a}: Reconstruction of the Lorenz-63 from Gaussian and ordinal observations, with $20\%$ of data removed at random times chosen independently for each modality. 
    \textbf{b}: Using only the Gaussian expert, 
    the corresponding ordinal observations can be decoded almost perfectly, including at times missing in the ordinal training data. Dashed lines indicate sections with missing data in \textbf{a} and \textbf{b}.}
    \label{fig:cross_modal_missing}
\end{figure}

Finally, MTF can easily handle missing data in one or more modalities, by simply dropping missing data points from the loss terms and using encoders like a mixture-of-experts (Appx. \ref{appx:mixture_of_experts}), as illustrated in Fig. \ref{fig:cross_modal_missing}. As Fig. \ref{fig:cross_modal_missing} highlights as well, MTF can also be employed for \textit{cross-modal inference}, predicting outcomes for unobserved modalities through the joint encoder model.

\subsection{DSR from Discrete Random Variables}\label{sec:discrete_dsr}

\begin{figure*}[!htb]
    \centering
	\includegraphics[width=0.96\linewidth]{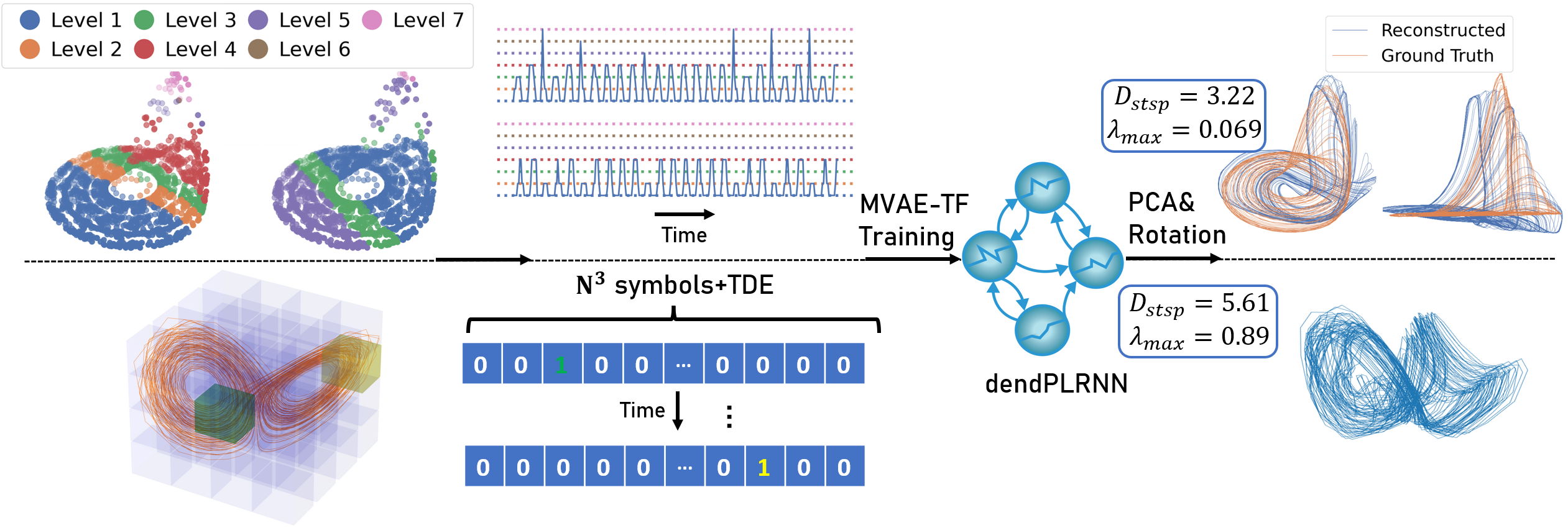}
	\caption{DS reconstruction from discrete observations by \ourmethodname\ ($M=30, B=15, K=30, \tau=10$). Top: Reconstruction of Rössler attractor from only ordinal time series. Bottom: DS reconstruction from symbolic coding of Lorenz attractor (see Fig. \ref{fig:symbolic_lorenz_categories} for true and predicted class label probabilities). Ground truth systems and their ordinal/symbolic encoding are on the left, corresponding reconstructions on the right. Note that in both cases the topology and general geometry are preserved, and maximal Lyapunov exponents closely match those of the true systems
 (Rössler: $\lambda_{\text{max}}^{\text{true}} \approx 0.072$, Lorenz: $\lambda_{\text{max}}^{\text{true}} \approx 0.903$). 
 TDE = temporal delay embedding.}
    \label{fig:discrete-DSR}
    \end{figure*}

Motivated by these results, we pushed the system even further and attempted DS reconstruction \textit{solely from ordinal data} (created as above), i.e. completely omitting continuous observations. This is profoundly more challenging than the multimodal setting with Gaussian data, since the ordinal process considerably coarse-grains the underlying continuous dynamical process. 
Since in this case we do not have a direct linear mapping between ground truth state space and that of the trained dendPLRNN (which in the case of Gaussian observations would simply be given by the linear decoder, \Eqref{eq:plrnn_obs}), we construct one post-hoc by optimizing a linear operator given by a linear dimensionality reduction (PCA) concatenated with a geometry-preserving rotation operation (see Methods \ref{sec:method:geometric}). Fig. \ref{fig:discrete-DSR} (top) confirms that, using \ourmethodname, successful DS reconstruction is still possible under these conditions (see also Fig. \ref{fig:lorenz96} for a higher-dimensional example of reconstruction from only ordinal data). 

Finally, we asked whether DS reconstruction might be feasible based on a purely \textit{symbolic representation} of the dynamics. For this, $4^3$ symbols were used corresponding to subregions defined by a $4 \times 4 \times 4$ grid superimposed on the attractor (this symbolic code was reduced further and then delay-embedded, 
see Appx. \ref{sec:methods:datasets} for full details). 
Fig. \ref{fig:discrete-DSR} (bottom) illustrates the general procedure, and also shows that, using \ourmethodname, successful DS reconstruction is -- in principle -- possible from just a symbolic (categorical) coding 
of the underlying chaotic attractor. This is to our knowledge the first time such a result has been shown. Comparable results could not be achieved by the multimodal SVAE or MS (see Table \ref{table_discrete}). Fig. \ref{fig:symbolic_lorenz_categories}\textbf{b} shows that even the maximum Lyapunov exponent $\lambda_{max}$ of the system reconstructed with \ourmethodname\ from just the symbolic coding of the Lorenz attractor often comes close to the true value (see also Fig. \ref{fig:discrete_supplement}).

\subsection{DSR from Neuroimaging and Behavioral Data}\label{sec:exp:fmri}

We evaluated \ourmethodname's performance on real multimodal time series using a dataset of functional magnetic resonance imaging (fMRI) recordings from human subjects undertaking various cognitive tasks in an fMRI scanner \citep{koppe_temporal_2014}. The data consisted of continuous blood-oxygenation-level-dependent (BOLD) time series acquired from 20 subjects at a sampling rate of $1/3$ \si{\hertz}. The first principal component of BOLD activity from each of 20 different brain regions was selected for analysis. The individual time series were rather short ($T=360$ per subject), posing a significant challenge for DS reconstruction. In addition to continuous BOLD activity, categorical time series of cognitive stage labels were given for each subject, corresponding to the five cognitive tasks that each subject went through repeatedly during the experiment (`Rest’, `Instruction’, `Choice Reaction Task [CRT]’, `Continuous Delayed Response Task [CDRT]’ and `Continuous Matching Task [CMT]’)\footnote{Note that CRT, CDRT, and CMT did not differ in terms of presented stimuli or response types, but only in cognitive load, thus reflecting processes internal to the subject.}, which we accommodated via a multi-categorical decoder model, \Eqref{eq:supp:cat_obs} in Methods \ref{sec:methods:obsmodels}.

\begin{table}
\caption{Comparison among multi-modal reconstruction methods for experimental fMRI+behavioral data. For each subject and training method, medians across $15$ trained models were first obtained for each measure, which were then averaged across $20$ subjects ($\pm$ SEM). SEM = standard error of the mean. X = value not accessible 
for this method. For 
abbreviations see Table \ref{table_noisy}. \\}
\centering
\renewcommand{\arraystretch}{1.1}
\scalebox{0.68}{
\begin{tabular}{ l c c c c}
\hline
Dataset& Method  & $D_{stsp}$ $\downarrow$ & $D_{H}$ $\downarrow$ & PE $\downarrow$  \\ \hline  
\multirow{5}{*}{fMRI}  & \ourmethodname\ & $\mathbf{0.55 \pm 0.04}$ & $\mathbf{0.301   \pm 0.007}$  &  $\mathbf{1.21  \pm 0.08}$ \\ 
                         & SVAE   & $1.9 \pm 0.22$   & $0.441 \pm 0.019$   &  $2.34 \pm 0.12$  \\
                        & BPTT  & $3.31 \pm 0.8$   & $0.52 \pm 0.05$   &  $2.8 \pm 0.15$  \\
                        & MS     & $1.06 \pm 0.14$   &  $0.373 \pm 0.012$               &  X  \\
                         & GVAE-TF& $0.67 \pm 0.06$   & $0.335 \pm 0.011$   &  $1.64 \pm 0.07$  \\
                         & BPTT-TF& $0.63 \pm 0.03$   & $0.312 \pm 0.006$   &  $1.39 \pm 0.05$  \\  
                           \hline
\end{tabular}
}
\label{table_fmri}
\end{table}

   \begin{figure*}[!htb]
    \centering
	\includegraphics[width=0.93\linewidth]{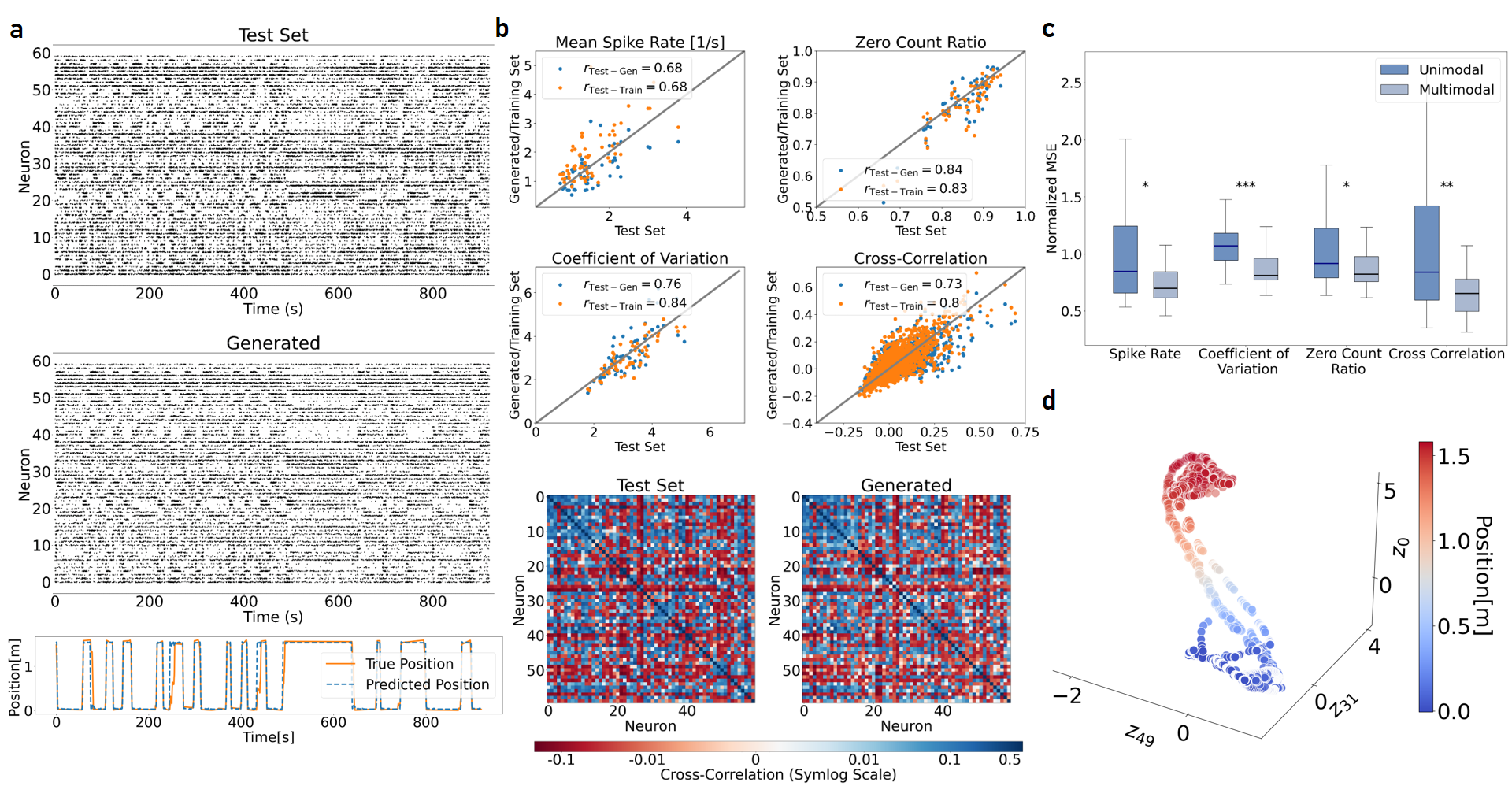}
	\caption{\textbf{a}: Example reconstructions of spike trains and spatial location of animal. Top: Spike train data from a test set not used for model training (topmost), and model-generated spike trains (below) 
 simulated from a data-inferred initial condition. Bottom: True and model-predicted spatial position on the test set. \textbf{b}, top: Correlation of various spike train statistics between test set and model-generated data (blue), and -- for comparison -- between experimental training and test set data (orange)
    : mean spike rate, zero count ratio, coefficient of variation, and correlation between cross-correlation coefficients. Diagonal gray lines are bisectrices, not regression lines. Bottom: Cross-correlation matrices among all neurons for the experimental test set (left) and model-generated spike data (right). \textbf{c}: Joint DSR 
    from both spatial and neural data significantly improves reconstructions compared to just neural data alone ($* \, p<0.05$, $*** \, p<0.001$). \textbf{d}: DSR model latent space (shown is a subspace), illustrating how the latent dynamics is organized according to the animal's spatial position (color-coded).
}
    \label{fig:hippocampus_main}
    \end{figure*}

This is the same dataset and setup as considered in \citet{kramer22a}, which we use here for comparability, repeating the same type of analyses as performed in that previous study for our \ourmethodname. We first compared the performance of a dendPLRNN trained with \ourmethodname\ only on the BOLD signals to the performance when the same algorithm was additionally provided with the categorical information, optimizing a joint Gaussian and categorical likelihood (see Appx. \Eqref{eq:supp:cat_obs}). 
As shown in Fig. \ref{fig:fmri_main}\textbf{a} and statistically confirmed by a paired t-test, according to $D_{stsp}$ ($t_{19}=2.45, p<.013$) and $D_{H}$ ($t_{19}=2.72, p<.007$), training the dendPLRNN with multimodal data significantly improved DS reconstruction compared to training on just the BOLD signals.
These results confirm that even in this empirical situation categorical data can significantly help to improve DS reconstruction, as \citet{kramer22a} observed previously for their algorithm.

Fig. \ref{fig:fmri_main}\textbf{b} further shows an example where on a left-out test set simulated BOLD activity and model-generated task labels closely resembled those of the real subject. 
Fig. \ref{fig:fmri_main}\textbf{c} investigates to what extent cross-modal links have been built by the trained system in its latent space by illustrating that after multimodal training, freely sampled latent activity trajectories from the dendPLRNN reflect the structure of the different tasks stages. 
Fig. \ref{fig:fmri_freely_gen} in Appx. \ref{sec:supp:fmri} provides generated time series for several example subjects. Table \ref{table_fmri}, finally, compares the DSR performance on the fMRI and behavioral data for all the methods introduced in Sect. \ref{sec:experiments}, with \ourmethodname\ substantially outperforming the multimodal SVAE from \citep{kramer22a}, again highlighting the crucial differences between these approaches.

\subsection{DSR from Neurophysiological and Location Data}\label{sec:exp:spikes}
As a second empirical test case, we used electrophysiological recordings of multiple single neurons from the CA1 region of the rodent hippocampus \citep{grosmark_dataset_2016}, a brain region involved in spatial navigation and memory \citep{buzsaki2006rhythms, moser_spatial_2017}. During the experiments the rats traversed a 1.6 m long track along which the longitudinal position of the animal was recorded, treated as a continuously valued Gaussian $1$d time series. Water rewards were provided at each of the two end points of the track, indicated to the model through brief external inputs. At the same time, for the data set on which we focused our analysis, the spiking activity (action potential time points) of $120$ neurons was recorded via intracortical electrodes (of which $60$ with reasonably high spiking rate were retained). 
Following common 
procedures in neuroscience \citep{zhou_learning_2020,schneider_learnable_2023}, spike trains were segmented into $200$ ms bins, resulting in a multivariate count time series (for details, see Methods \ref{sec:methods:hippo}). These were modeled with a negative-binomial decoder 
(\Eqref{eq:negative_binomial_obs}), capturing the high dispersion better than Poisson or zero-inflated Poisson models.

Multimodal (position and spike count) data were split into a training and a test set of $4600$ time steps each. Example true and model-generated spike trains for the 
test set are in Fig. \ref{fig:hippocampus_main}\textbf{a} (see Fig. \ref{fig:hippo_train} for training set).
Fig. \ref{fig:hippocampus_main}\textbf{b} compares various standard spike train statistics like the mean firing rate, mean zero rate, coefficient of variation, 
and the neural cross-correlation matrix between experimental and model-generated data, iterated forward from one inferred initial state across the whole test set.
The agreement in spiking statistics between model-simulated long-term behavior and test data was in fact as good as between different segments (training and test set) of the experimental data itself.
Moreover, the predicted position on the track closely matched that of the real behavior for the test set (Fig. \ref{fig:hippocampus_main}\textbf{a}). 
Using both data sources (spike trains and position) led to far better reconstructions across all spike train performance metrics, and overall more robust reconstructions, than when only spike count data were provided for model training (Mann-Whitney U-test across $50$ trained models, $p<0.025$ for all performance metrics; Fig. \ref{fig:hippocampus_main}\textbf{c}).
In the CA1 region, place cells are known to encode spatial information  \citep{okeefe_hippocampus_1978}. Our results indicate that the MTF algorithm can leverage this information to build a joint latent model of the neural recordings and movement of the rat (Fig. \ref{fig:hippocampus_main}\textbf{d}). 

\section{Conclusions}\label{sec:conclusion}
Here we introduced a novel training framework (MTF) for DSR from non-Gaussian and multimodal time series data. While DS reconstruction is meanwhile a large field in scientific machine learning 
\citep{brunton_discovering_2016, pathak_model-free_2018, lejarza_data-driven_2022, brenner22a, pmlr-v202-hess23a, chen2023deep}, reconstruction based on discrete or multimodal data has hardly been addressed so far, although such scenarios are commonplace in many areas like medicine, neuroscience, or climate research. Here we utilize recent insights on guiding the training process by control signals, as in STF \citep{mikhaeil_difficulty_2022} and GTF \citep{pmlr-v202-hess23a}, within a novel multimodal DSR framework. In our approach, a sparse TF signal is generated by an MVAE that translates different data modalities into a common latent code in order to guide a DSR model. Our general framework is very flexible and allows to easily swap DSR, encoder, or decoder models for the problem at hand. For diverse benchmarks and sampling conditions, MTF clearly outperforms various other training setups and models. Moreover, we found MTF to be quite robust performance-wise, and not very sensitive to precise hyper-parameter tuning. 

We further highlighted on examples from neuroscience how the DSR model's latent space exposes meaningful relations between different data modalities (neural activity and behavior in our case). In contrast to other recently proposed methods for construing a common latent space from multiple data sources \citep{schneider_learnable_2023}, our method actually delivers a \textit{generative model of the system dynamics}. A crucial difference to `classical' multimodal SVAEs \citep{kramer22a} is that our algorithm allows to optimally control trajectory and gradient flows whilst training, a key ingredient for successful DS reconstruction 
\citep{mikhaeil_difficulty_2022,pmlr-v202-hess23a,eisenmann2023bifurcations}. One potentially interesting future direction may be engineering ways of incorporating control signals directly at the encoder level.

This way, MTF was even able, under certain conditions, to recover chaotic attractors from ordinal or categorical observations alone, including sensible estimates of their maximum Lyapunov exponent.
This, to our knowledge, has never been shown before, and at first glance appears puzzling, as an ordinal or symbolic coding of the dynamics seems to remove much of the geometrical information.
However, results in the field of symbolic dynamics ensure that under certain conditions a symbolic encoding of a DS preserves topological invariants and Lyapunov spectra \citep{osipenko2000spectrum, hubler2012symbolic, mccullough2017regenerating,zhang2017constructing, lind2021introduction}, and delay embedding theorems have been formulated for non-continuous signals like point processes \citep{sauer1994reconstruction, sauer1995interspike}. While we foresee the major applications of MTF in truly multimodal settings, our observations here may also pave the way for interesting novel DSR applications to purely symbolic data. For instance, reconstructing dynamics from categorical consumer data or sequences of behavioral choices (on which classically reinforcement learning models would be trained) may become feasible, or even from language data \cite{elman_language_1995}, e.g. recovering the dynamics of underlying beliefs or values \cite{tsakalidis_overview_2022}.

While here we presented first results on DS reconstruction for discrete data, whether and how much of the original state space topology of a data-generating DS can be recovered from non-continuous, non-Gaussian random variables remains an important topic for future theoretical and empirical research. Another open question remains under which conditions different modalities support (or interfere with) each other in the DSR process.

\section*{Software and Data}
All code created is available at \url{https://github.com/DurstewitzLab/MTF}.

\section*{Acknowledgements}
This work was funded by the European Union’s Horizon 2020 Programme under grant agreement 945263 (IMMERSE), by the German Research Foundation (DFG)
within the collaborative research center TRR-265 (project A06) , by the German Research Foundation (DFG) within and Germany’s Excellence Strategy EXC 2181/1 – 390900948 (STRUCTURES), and by a living lab grant by the Federal Ministry of Science, Education and Culture (MWK) of the state of 
Baden-Württemberg, Germany (grant number 31-7547.223-7/3/2), to DD and GK.

\section*{Impact Statement}

This paper presents a novel approach in the context of scientific machine learning, with applications in fields like neuroscience. We believe the primary ethical implications of this approach are positive. While we can not specifically highlight any ethical concerns, given the broad scope of possible applications of our approach, potential misuses can not be a priori excluded.

\bibliography{MTF}
\bibliographystyle{icml2024}

\newpage
\appendix
\onecolumn

\setcounter{figure}{0} 
\renewcommand{\thefigure}{A\arabic{figure}}
\renewcommand{\theHfigure}{A\arabic{figure}}
\setcounter{table}{0}
\renewcommand{\thetable}{B\arabic{table}}

\section{Methods}\label{sec:method}

\subsection{MTF framework}\label{sec:method:mtf}
The MTF framework consists of four major components, 1) a DSR model for approximating the flow of the underlying, observed DS, 2) a set of modality-specific decoder models that capture the statistical particularities of the observed time series, 3) an MVAE for integrating all different data modalities in order to produce a joint latent code which can be used for TF, and 4) the actual MTF training algorithm. Here we describe each of these components in turn.

\paragraph{DSR model} While our framework is generic and works with any type of DSR model, here we used the dendPLRNN, 
\Eqref{eq:plrnn_lat}, recently introduced specifically for DSR \citep{brenner22a} (see Appx. Table \ref{table_rnns} for comparison with other RNN models). A major advantage of the dendPLRNN and related formulations is that it provides some degree of analytical tractability, which is of major importance in scientific settings where we are usually interested in analyzing the data-inferred model further as a formal surrogate for the real system. For example, for the dendPLRNN, we have efficient algorithms for precisely locating all fixed points, cycles, and bifurcation manifolds \citep{eisenmann2023bifurcations}, and it allows for translation into an equivalent continuous-time representation \citep{monfared_transformation_2020}, which further eases certain types of DS analysis.

\paragraph{Decoder models}\label{sec:methods:obsmodels}
To infer the latent process dendPLRNN jointly from multiple data modalities, it is connected to different decoder models that take the distinct distributional properties of each modality into account, Eqs. \ref{eq:gauss_obs_model}-\ref{eq:poisson_obs_model}.

For \textit{normally distributed} data, the decoder may take the simple linear Gaussian form
\begin{equation}\label{eq:plrnn_obs}
	\bm{x}_t = \bm{B} \bm{z}_t + \bm{\eta}_t,
\end{equation}
with factor loading matrix $\bm{B} \in \mathbb{R}^{N \times M}$, and Gaussian observation noise $\bm{\eta}_t \sim\mathcal{N}(\bm{0},\bm{\Gamma})$ with covariance $\bm{\Gamma} \in \mathbb{R}^{N \times N}$ (assumed to be diagonal here).

\textit{Ordinal data}, in contrast, are not associated with a metric space, but there is a natural ordering between variables, as e.g. in survey data in economy or psychology commonly assessed through Likert scales \citep{likert_technique_1932}. Treating ordinal data as metric can lead to a variety of problems, as pointed out in \citep{liddell_analyzing_2018}. Ordinal observations are coupled to latent states via a generalized linear model \citep{mccullagh_regression_1980}. Here, specifically, we assume that the ordinal observations $\bm{o}_{t}$ are derived from an underlying unobserved continuous variable $u_{it}$, which is linked to the latent states $\bm{z}_t$ via a linear model
\begin{equation}
  u_{i t}=\boldsymbol{\beta}_i^T \bm{z}_t+\epsilon_{i t},
\end{equation}
 where $\bm{\beta}_i^T \in \mathbb{R}^M$ are the model parameters and $\epsilon_{i t}$ is an independently distributed noise term. The distributional assumptions about the noise term $\epsilon_{i t}$ determine which link function to use. A Gaussian assumption leads to an ordered probit model, while a logistic assumption leads to an ordered logit model \citep{winship_regression_1984}. While both models lead to similar results, we found the ordered logit model to work slightly better in practice, and hence we focus on it here.
Inverting the link function leads to an expression for the cumulative probabilities:
\begin{equation}\label{eq:ord_obs}
    p\left(o_{i t} \leq k | \bm{z}_t\right)=\frac{\exp \left(\beta_{i k}^0-\boldsymbol{\beta}_i^T \bm{z}_t\right)}{1+\exp \left(\beta_{i k}^0-\boldsymbol{\beta}_i^T \bm{z}_t\right)} .
\end{equation}
The probability masses $p\left(o_{i t}=k | \bm{z}_t\right)$ follow from the cumulative distribution via $p\left(o_{i t}=k | \bm{z}_t\right)=p\left(o_{i t} \leq k | \bm{z}_t\right)-p\left(o_{i t} \leq k-1 | \bm{z}_t\right)$,
from which we can compute the log-likelihood as 
\begin{equation}\label{eq:log_likelihood_ordinal}
\log p_{\bm{\theta}}(\bm{O} | \bm{Z})=\sum_i^N \sum_t^T \sum_k^K\left[o_{i t}=k\right] \log p\left(o_{i t}=k | \bm{z}_t\right) .
\end{equation}

\textit{Categorical observations} are, like ordinal observations, not associated with a metric space, and in contrast to ordinal data, there is also no natural ordering between variable values. To couple categorical observations to the latent states, we employed the natural link function given by
\begin{align} \label{eq:supp:cat_obs}
&\pi_i=\frac{\exp \left(\boldsymbol{\beta}_i^T \mathbf{z}_t\right)}{1+\sum_{j=1}^{K-1} \exp \left(\boldsymbol{\beta}_j^T \mathbf{z}_t\right)} \quad \forall i \in\{1 \ldots K-1\}  \\
&\pi_K=\frac{1}{1+\sum_{j=1}^{K-1} \exp \left(\boldsymbol{\beta}_j^T \mathbf{z}_t\right)} \nonumber
\end{align}
Here, the parameters $\boldsymbol{\beta}_i \in \mathbb{R}^{M \times 1}$ constitute the respective regression weights for category $i=1 \ldots K-1$, with the total probability over all categories $\sum_{i=1}^K \pi_i=1$.
This leads to the following log-likelihood for the categories:
\begin{align} \label{eq:log_likelihood_categorical}
\log p_{\bm{\theta}}(\bm{O} | \bm{Z}) \nonumber \
= \sum_i^N \sum_t^T \sum_k^K \left[o_{i t}=k\right] \log \left( \frac{\exp \left(\bm{\beta}_k^T \bm{z}t\right)}{1+\sum_{j=1}^{K-1} \exp \left(\bm{\beta}_j^T \bm{z}_t\right)} \right)
\end{align}
where $\left[o_{i t}=k\right]$ is an indicator function that is $1$ if the observation $o_{it}$ belongs to category $k$ and $0$ otherwise.

For \textit{count observations} $\left\{\bm{c}_t\right\}_{t=1}^T$, with $\bm{c}_t=\left(c_{1 t}, \dots, c_{L t}\right)^T$, we tried three different decoder models. First, a standard Poisson model,
\begin{equation} \label{eq:poisson_obs}
\quad p_{\bm{\theta}}\left(c_{l t} | \bm{z}_t\right)= \frac{\lambda_{l t}^{c_{l t}}}{c_{l t} !} e^{-\lambda_{l t}} .
\end{equation}
These probabilities are related to the latent states via the log-link function, $\log \lambda_{l t}=\gamma_0^{(l)}+\sum_{m=1}^M \gamma_m^{(l)} z_{m t}$, where $\boldsymbol\gamma^{(l)}$ is a vector of coefficients.
Thus, $\lambda_{l t}=e^{\gamma_0^{(l)}+\boldsymbol\gamma^{(l)} \bm{z_t}}$ is the expected count for the $l^\text{th}$ observation at time $t$. The total log-likelihood for observed counts $\bm{C}$ is then given by: 
\begin{equation} \label{eq:log_likelihood_counts}
\log p_{\bm{\theta}}(\bm{C} | \bm{Z}) = \sum_{t=1}^T \sum_{l=1}^L \left[ c_{l t} \log \lambda_{l t} - \lambda_{l t} - \log(c_{l t}!) \right],
\end{equation}
Alternatively, we tested a zero-inflated Poisson (ZIP) observation model \cite{lambert_zero-inflated_1992}. The ZIP model is designed for count data that has an excess of zero counts compared to what would be expected from a traditional Poisson distribution, and thus can naturally handle overdispersion often observed in real data, such as the spike trains displayed in Fig. \ref{fig:hippocampus_main}. The ZIP model addresses this by combining a binary process that determines whether a count is zero (with probability $\pi_{lt}$) with a Poisson distribution. The combined probability of an observed count given both processes is then given by
\begin{equation} 
\quad p_{\bm{\theta}}\left(c_{l t} \mid \bm{z}_t\right)=\left\{\begin{array}{cl}
\pi_{l t}+\left(1-\pi_{l t}\right) e^{-\lambda_{l t}} & \text { if } c_{l t}=0 \\
\left(1-\pi_{l t}\right) \frac{\lambda_{l t}^{c_{l t}}}{c_{l t} !} e^{-\lambda_{l t}} & \text { if } c_{l t}>0
\end{array}\right.
\end{equation}
The probabilities are connected to the latent states via a logit and a log-link function $\log \frac{\pi_{l t}}{1-\pi_{l t}}=\beta_0^{(l)}+\sum_{m=1}^M \beta_m^{(l)} z_{m t}$ and $\quad \log \lambda_{l t}=\gamma_0^{(l)}+\sum_{m=1}^M \gamma_m^{(l)} z_{m t}$, where $\gamma^{(l)}$ and $\beta^{(l)}$are coefficient vectors.
Thus,$\quad \lambda_{l t}=e^{\gamma^{(l)} z_t}$ is the expected count for the $l^\text{th}$ observation and $\pi_{l t}=\frac{e^{\beta^{(l)} z_t}}{1+e^{\beta^{(l)} z_t}}$ is the probability of observing a zero.

Finally, we tested a negative binomial model given by
\begin{equation} \label{eq:negative_binomial_obs}
p_{\bm{\theta}}\left(c_{l t} | \bm{z}_t\right)= \frac{\Gamma(c_{l t} + \phi_l)}{\Gamma(\phi_l) c_{l t}!} \left(\frac{\phi_l}{\mu_{l t} + \phi_l}\right)^{\phi_l} \left(\frac{\mu_{l t}}{\mu_{l t} + \phi_l}\right)^{c_{l t}}, 
\end{equation}
where $\mu_{l t}$ is the mean count and $\phi_l$ the dispersion parameter of the negative binomial distribution for the $l^\text{th}$ observation at time $t$. As for the Poisson model, we 
use a log-link function, $\log \mu_{l t}=\gamma_0^{(l)} + \sum_{m=1}^M \gamma_m^{(l)} z_{m t}$, with $\boldsymbol\gamma^{(l)}$ a vector of coefficients, and $z_{m t}$ the $m^\text{th}$ latent variable at time $t$. Properly accounting for dispersion significantly improved the modeling of the spike counts. For the evaluation we used the the likelihood function implemented in the \texttt{torch.distributions.NegativeBinomial} class from the PyTorch library.

\paragraph{Multimodal variational autoencoder}\label{sec:methods:mvae}

The MVAE consists of a set of decoder models, $p_{\bm{\theta}}(\bm{Y}|\bm{\tilde{Z}})$, shared with the DSR model, with examples given in Eqs. \ref{eq:mvae_gauss_obs_model}-\ref{eq:mvae_poisson_obs_model} and 
above, a prior model $p_{\bm{\theta}}(\bm{\tilde{Z}})$ instantiated by the DSR model, see \Eqref{eq:prior_term}, and the encoder (approximate posterior) $q_{\bm{\phi}}(\bm{\tilde{Z}}|\bm{Y})$, where a sample from this distribution provides the TF signal (see also Fig. \ref{fig:multimodalteacherforcing}). This setup is modular and all model components may be flexibly chosen. 
We assume conditional independence (given latent states) of the observations, such that the likelihood terms related to the observations simply sum up, i.e. 
$\log p_{\bm{\theta}}(\bm{{Y}}|\bm{\tilde{Z}})=\sum_{t=1}^T\left(\log p_{\bm{\theta}}(\bm{x}_t|\bm{\tilde{z}}_t)+\log p_{\bm{\theta}}(\bm{o}_t|\bm{\tilde{z}}_t)+\log p_{\bm{\theta}}(\bm{{c}_t}|\bm{\tilde{z}}_t)\right)$ for the decoders specified in Eqs. \ref{eq:mvae_gauss_obs_model}-\ref{eq:mvae_poisson_obs_model} (with parameters shared with Eqs. \ref{eq:gauss_obs_model}-\ref{eq:poisson_obs_model}).

For the encoder part, as the latent dendPLRNN process itself is conditionally Gaussian, we make a Gaussian assumption for the variational density $q_{\bm{\phi}}(\bm{\tilde{Z}}|\bm{Y})=\mathcal{N}(\bm{\mu}_{\bm{\phi}}(\bm{Y}),\bm{\Sigma}_{\bm{\phi}}(\bm{Y}))$ to approximate the true 
posterior $p(\bm{\tilde{Z}}|\bm{Y})$, where mean and covariance are functions of the data. We further use the common mean field approximation to factorize $q_{\bm{\phi}}(\bm{\tilde{Z}}|\bm{Y})$ across time \citep{girin_dynamical_2021}. The mean $\bm{\mu}_{\theta}(\bm{Y})$ and covariance $\bm{\Sigma}_{\theta}(\bm{Y})$ of the approximate posterior are parameterized by one-dimensional (across time) convolutional layers \citep{brenner22a}, using observations within the window defined by the input kernel. For the mean, we use a 4-layer stack of convolutional layers with decreasing kernel sizes \( 11, 8, 5, 3 \), while the diagonal covariance was parameterized by a single convolutional layer with a kernel size of \( 11 \). We also tested various other choices for the encoder, as discussed in Appx. \ref{sec:appx:encoders}.

\begin{figure}[!htb]
    
    \centering
	\includegraphics[width=0.99\linewidth]{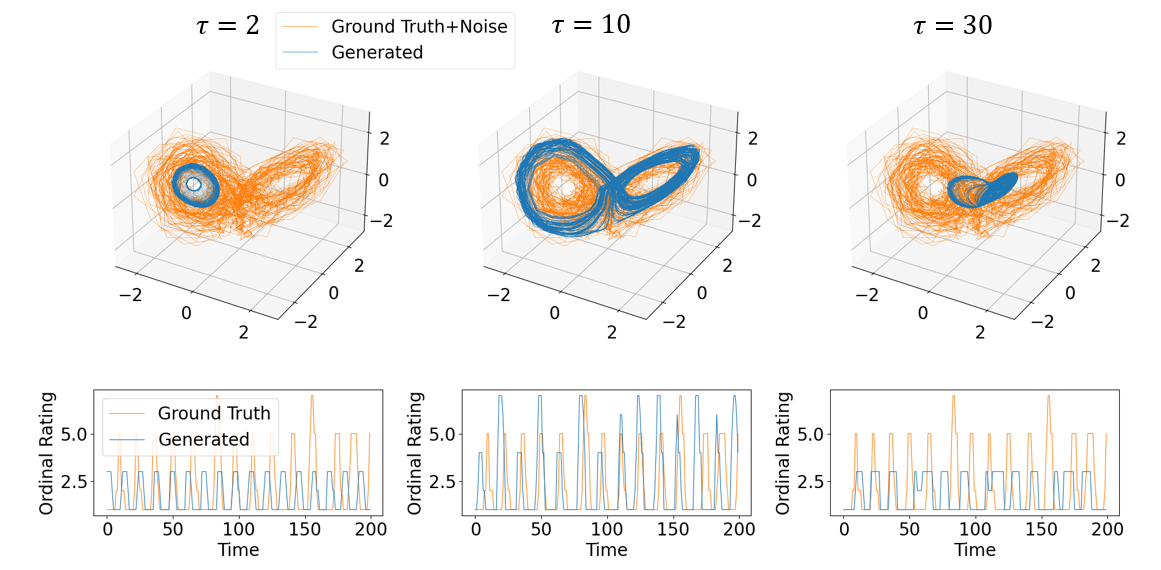}
 \caption{Example reconstructions (best out of $10$ trained models each) of a dendPLRNN trained via \ourmethodname, compared to highly noisy Gaussian and ordinal training data (orange) produced by the Lorenz system. Reconstruction quality crucially depends on the choice of an optimal forcing interval, similar as observed for 
 STF in \citet{mikhaeil_difficulty_2022}. Choosing the interval too small (left) or too large (right) leads to significantly worse reconstructions.
 }
 \label{fig:influence_tau}
\end{figure}

\paragraph{Multimodal teacher forcing} 
Once we have a mechanism for generating a multimodal TF signal, our framework can accommodate any form of TF training scheme, like STF \citep{mikhaeil_difficulty_2022} or GTF \citep{pmlr-v202-hess23a}. 
Consider an observed (multimodal) time series $\bm{Y}=\{\bm{y}_1,\bm{y}_2,\dots,\bm{y}_T\}$ generated by a DS we would like to reconstruct. From this we create a control series $\hat{\bm{Z}}=\{\hat{\bm{z}}_1,\hat{\bm{z}}_2,\dots,\hat{\bm{z}}_T\}$, $\hat{\bm{z}}_t\in \mathbb{R}^{M}$. In STF or GTF, one obtains $\hat{\bm{Z}}$ from inverting the decoder model, projecting observations into the DSR model's latent space. Here, in contrast, we equate the control series with the latent code $\bm{\tilde{Z}}=\{\bm{\tilde{z}}_1,\dots,\bm{\tilde{z}}_T\}$, $\bm{\tilde{z}}_t\in \mathbb{R}^{K}$, produced by the MVAE. 
For the present evaluation we focused on a specific form of STF \citep{brenner22a}, where the first $K$ latent states are replaced by the control states $\hat z_{k,l\tau+1} = \tilde z_{k,l\tau+1},\  k\leq K$, at times $l\tau+1,\ l \in \mathbb{N}_0$, with forcing interval $\tau \geq 1$, while the remaining latent states, $\hat z_{k,l\tau+1} = z_{k,l\tau+1}, \ k> K$, remain unaffected. 
Defining $\mathcal{F}=\{l\tau+1\}_{l \in \mathbb{N}_0}$, the dendPLRNN updates can thus be written as
\begin{align}\label{eq:forcing}
\bm{z}_{t+1} =
  \begin{cases}
       \text{\ournetworkname}(\bm{\hat z}_{t}) & \text{if $t\in \mathcal{F}$} \\
       \text{\ournetworkname}(\bm{z}_{t}) & \text{else} 
  \end{cases}
\end{align}
For the initial condition $\bm{z}_1$ of the latent process, if $K<M$, the $M-K$ remaining states are randomly sampled from a standard normal distribution. As shown in \citet{mikhaeil_difficulty_2022}, the best tradeoff between exploding gradients and capturing relevant long-term dependencies is achieved when choosing the forcing interval $\tau$ according to the system's maximal Lyapunov exponent (predictability time). However, for non-continuous data (like counts), this cannot readily be determined, in which case the forcing interval $\tau$ may be regarded as a hyper-parameter determined via grid search (see also Appx. Fig. \ref{fig:influence_tau} for example reconstructions for different choices of $\tau$). 

To ensure that the dendPLRNN and MVAE can share all decoder model parameters, only the first $K \leq M$ states of the 
generated latent trajectory $\bm{Z}$ (using the generated states prior to forcing) 
are then used to compute the modality-specific negative log-likelihoods for the observed multimodal time series,
\begin{equation}\label{loss_PLRNN}
\mathcal{L}_\text{DSR}=-\sum_{t=1}^T(\log p_{\bm{\theta}}(\bm{{x}_t}|\bm{z}_{1:K,t})+\log p_{\bm{\theta}}(\bm{o}_t|\bm{z}_{1:K,t})+\log p_{\bm{\theta}}(\bm{c}_t|\bm{z}_{1:K,t})),
\end{equation}
 using the decoder models from Eqs. \ref{eq:gauss_obs_model}-\ref{eq:poisson_obs_model}. 
 
Likewise, to connect the latent codes of the MVAE and the RNN, the model prior of the MVAE is instantiated through the RNN by taking $\bm{\mu}_t=\bm{z}_{1:K,t}$, i.e., only the first $K$ states of the generated latent sequence $\{\bm{z}_t\}$. As the initial state $\bm{z}_1$ is estimated directly from the encoded state $\bm{\tilde{z}}_1$, the term for $t=1$ evaluates to zero. Setting $L=1$, the \textit{consistency loss}, \Eqref{eq:prior_term}, between encoded and generated latent state paths thus comes down to
\begin{equation} \label{eq:latent_loss}
\mathcal{L}_\text{con}=\frac{1}{2}\sum_{t=2}^T\bigl(\log|\bm{\Sigma}| 
+  (\bm{\tilde{z}}_{t}-\bm{z}_{1:K,t})^{\top}\bm{\Sigma}^{-1}(\bm{\tilde{z}}_{t}-\bm{z}_{1:K,t})\bigl).
\end{equation}

The total \ourmethodname\ loss is hence given by the 
loss in \Eqref{eq:latent_loss} that ensures consistency between the latent codes of the MVAE 
and DSR model, 
the DSR model loss from the likelihoods of the observed time series $\bm{Y}$ given the predicted latent path $\bm{Z}$ (\Eqref{loss_PLRNN}), and the remaining terms from the ELBO, \Eqref{eq:elbo} (noting that $\mathcal{L}_\text{con}=-\mathbb{E}_{q_{\bm{\phi}}}[\log p_{\bm{\theta}}(\bm{\tilde{Z}})]$):
\begin{equation}
   \mathcal{L}_\text{MTF}= -\mathbb{E}_{q_{\bm{\phi}}}[\log p_{\bm{\theta}}(\bm{Y}|\bm{\tilde{Z}})] -\mathbb{H}_{q_{\bm{\phi}}}(\bm{\tilde{Z}} | \bm{Y}) +\mathcal{L}_\text{con}+\mathcal{L}_\text{DSR}
\end{equation}
To train the dendPLRNN with \ourmethodname, RAdam \citep{liu_variance_2020} was used with a learning rate scheduler that iteratively reduced the learning rate from $10^{-3}$ to $10^{-5}$ during training. For each epoch, we randomly sampled sequences of length $T_{seq}=300$ from the total training data with a batch size of $16$, except for the fMRI data, where we chose $T_{seq}=72$ due to the short overall length of the data ($T=360$ per subject). The network weights $\bm{A}$, $\bm{W}$ and $\bm{h}$ from \Eqref{eq:plrnn_lat} were initialized as suggested in \citet{talathi_improving_2016}. Expansion term thresholds $\{\bm{h}_b\}$ and input matrix entries $[\bm{U}]_{ij}$ were drawn from a standard normal distribution, while dendritic slopes $\alpha_b \sim \mathcal{U}(-B^{-\frac{1}{2}}, B^{-\frac{1}{2}})$ were initialized uniformly.

\subsection{Performance measures}\label{sec:exp:measures}

\paragraph{Geometrical measure}
To assess the (dis-)agreement $D_{\textrm{stsp}}$ between the data distribution $p_{\text {true}}(\vx)$ and the model-generated distribution $p_{\text {gen}}(\vx|\vz)$ across \textit{state space}, $\hat{p}_{\text {true}}(\vx)$ and $\hat{p}_{\text {gen}}(\vx | \vz)$ are estimated by sampling $100$ trajectories with randomly drawn initial conditions and $1000$ time steps each. Transients are removed from each sampled trajectory to ensure that the (occupation) measure is evaluated on the limit set. The match between distributions is then approximated by binning the state space into discrete bins \citep{koppe_identifying_2019}: 
\begin{align}\label{eq:D_stsp} 
D_{\mathrm{stsp}}\left(p_{\mathrm {true }}(\vx), p_{\mathrm {gen }}(\vx | \vz)\right) \approx \sum_{k=1}^{K} \hat{p}_{\mathrm {true }}^{(k)}(\vx) \log \left(\frac{\hat{p}_{\mathrm{true }}^{(k)}(\vx)}{\hat{p}_{\mathrm {gen }}^{(k)}(\vx | \vz)}\right).
\end{align}
Here, $K$ is the total number of bins.  A range of $2 \times$ the data standard deviation on each dimension was partitioned into $m$ bins, leading to a total of $K=m^{N}$ bins, where $N$ is the dimension of the ground truth system. 
Due to the exponential scaling of the number of bins with system dimensionality, for the $6$-dimensional Lewis-Glass network model and the fMRI data we instead used an approximation of $p_{\text {true}}(\vx)$ and $p_{\text {gen}}(\vx|\vz)$ based on Gaussian mixture models placed along trajectories, as described in \citet{brenner22a}.

\paragraph{Geometric reconstruction measure in the absence of continuous observations}
 \label{sec:method:geometric}

\begin{figure}[!htb]
    \centering
	\includegraphics[width=0.75\linewidth]{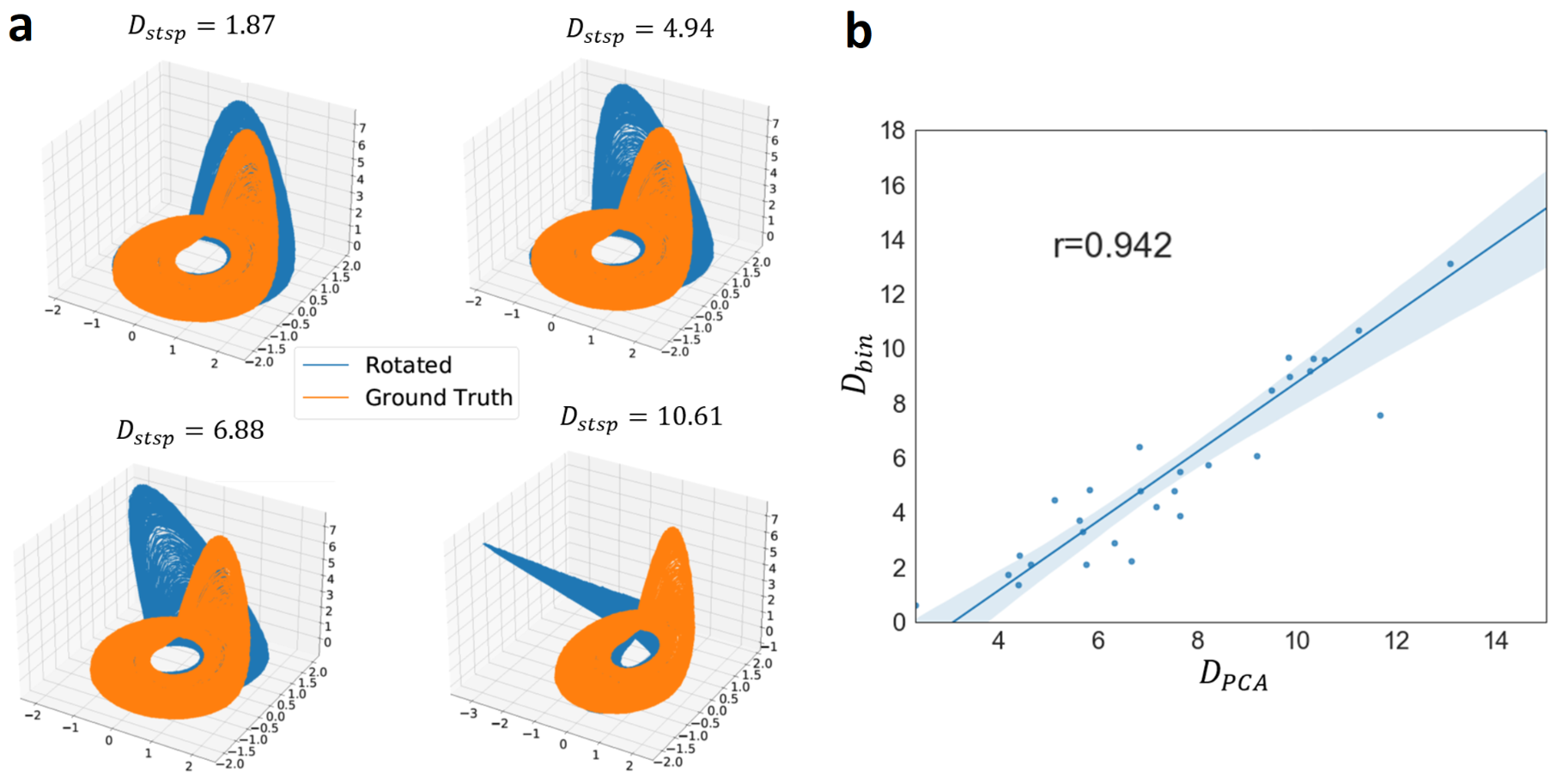}
	\caption{\textbf{a}:  Ground truth and rotated attractors of the Rössler system with associated $D_{stsp}$-values. \textbf{b}: Correlation between geometrical reconstruction measures for the Rössler system directly in observation space given a co-trained linear (Gaussian) observation model ($D_{bin}$), and from a 3d PCA projection of latent space followed by an optimal rotation of the reconstructed attractor ($D_{PCA}$), based on a total of
    $30$ trained models.}
    \label{fig:metrics_panel}
    \end{figure}

\begin{figure}[!htb]
    \centering
	\includegraphics[width=0.99\linewidth]{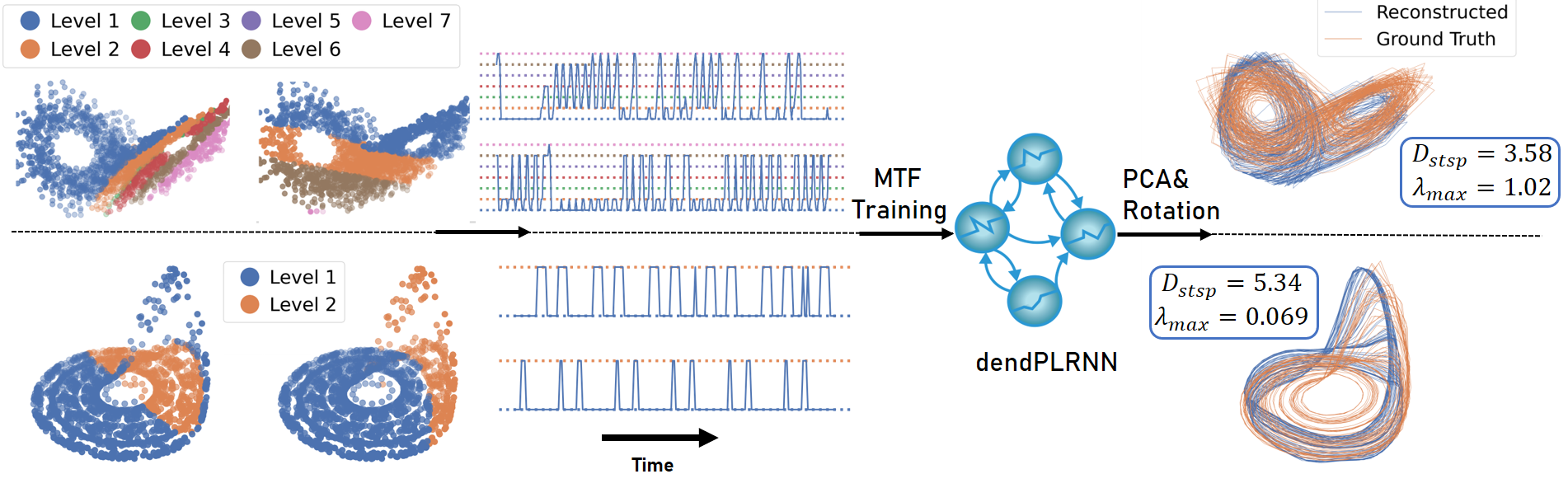}
	\caption{Top: DS reconstruction from eight ordinal observations with seven levels each produced by the Lorenz-63 system. Example traces of ordinal time series (center), and example reconstruction by \ourmethodname\ (right) 
    ($M=20, B=10, K=15, \tau=10$), preserving the butterfly geometry of the Lorenz attractor. Bottom: DS reconstruction of the Rössler attractor from from $75$ ordinal observations with two levels each.}
    \label{fig:discrete_supplement}
    \end{figure}

 If the underlying DS was observed only through time series of discrete random variables, we lack a direct mapping between the true and reconstructed continuous state spaces. In this case, to still assess to what degree reconstructed systems agree in terms of attractor geometry, we require a mapping between the two state spaces that does not introduce any additional degrees of freedom, but consists only of 1) a projection into a space of the same dimensionality (and re-standardization of variables) followed by 2) a (geometry-preserving) rotation. This was to ensure that the quality of geometrical agreement can be attributed solely to the reconstruction method and not to any post-hoc fitting.
For the first step, we used Principal Component Analysis (PCA) to reduce the dendPLRNN's latent space to the same dimensionality $N$ as that of the ground truth system (which usually is of lower dimensionality). Afterwards, all axes were re-standardized (as for the original system). In a second step, a rotation matrix was then determined to rotate the latent state space such as to minimize the same Kullback-Leibler measure $D_{stsp}$ that was used to assess agreement in attractor geometries, see  Fig. \ref{fig:metrics_panel} for an example. The optimal rotation matrix was determined by grid search over the space of rotation matrices, as we found numerical optimization to often yield inferior results. Note that this operation \textit{does not alter the geometry} of objects in the latent space but merely rotates them such that they are best aligned with their ground truth counterparts (we also attempted Procrustes analysis \citep{gower_generalized_1975}, using the Procrustes Python library \citep{Meng2022procrustes}, to determine the best affine mapping between spaces directly, but found this generally to be inferior despite actually being less conservative than our approach).
Comparing different grid- and step sizes in preliminary runs, we fixed parameters such that a single grid search takes no more than $30-60$ seconds on a single CPU. To confirm that this procedure yields results in agreement with those obtained from a co-trained linear-Gaussian model fed with continuous observations, we compared $D_{stsp}$ computed in observation space (`$D_{bin}$') as outlined above to $D_{stsp}$ computed on the corresponding dimensionality-reduced latent states using our PCA+rotation method (`$D_{PCA}$'). As shown in Fig. \ref{fig:metrics_panel}, these two measures were indeed highly correlated, $r\approx 0.94$. Example reconstructions from solely ordinal observations are given in Figs. \ref{fig:discrete-DSR} and \ref{fig:discrete_supplement}.
For the $6$-dimensional Lewis-Glass chaotic network model, performing a grid search over rotation matrices in the observation space was unfortunately no longer computationally feasible, such that in this case we resorted to Procrustes analysis (see above; generally, the Procrustes method aims to superimpose two data sets by optimally translating, rotating, and scaling them, preserving geometric similarity). In this case, the correlation between the $D_{stsp}$ measures obtained by a co-trained linear model and the one obtained post-hoc via the Procrustes-transformed space dropped to $r\approx 0.57$, but was still significant ($t_{28}=3.77, p<.001$).

\paragraph{Power spectrum Hellinger distance}
The power spectrum Hellinger distance ($D_{H}$) was obtained by first sampling a time series of $100,000$ time steps and computing dimension-wise Fast Fourier Transforms (using \texttt{scipy.fft}) for both the ground truth system and simulated time series, after discarding transients. The noise-dominated high-frequency tails of the spectra were cut off, and the power spectra were slightly smoothed with a Gaussian kernel and normalized.
We then computed the Hellinger distance \citep{mikhaeil_difficulty_2022} between smoothed power spectra of ground-truth, $F(\omega)$, and generated, $G(\omega)$, trajectories by
\begin{equation}
H(F(\omega), G(\omega))=\sqrt{1-\int_{-\infty}^{\infty} \sqrt{F(\omega) G(\omega)} d \omega} \in[0,1]
\end{equation}
The dimension-wise Hellinger distances were then averaged to yield the $D_H$ values reported.

\paragraph{Spearman autocorrelation function (SACF)}
To assess temporal agreement between model-generated and ground truth time series for \textit{ordinal} and \textit{count} observations, we computed a measure based on the average SACF, defined as:
\begin{equation} \label{eq:SACF}
SACF_i(\tau) = \frac{\sum_{t=1}^{T-\tau} (r_{i,t} - \overline{r}_i) \cdot (r_{i,t+\tau} - \overline{r}_i)}{\sum_{t=1}^T (r_{i,t} - \overline{r}_i)^2},
\end{equation}

\begin{figure}[!htb]
\centering
	\includegraphics[width=0.75\linewidth]{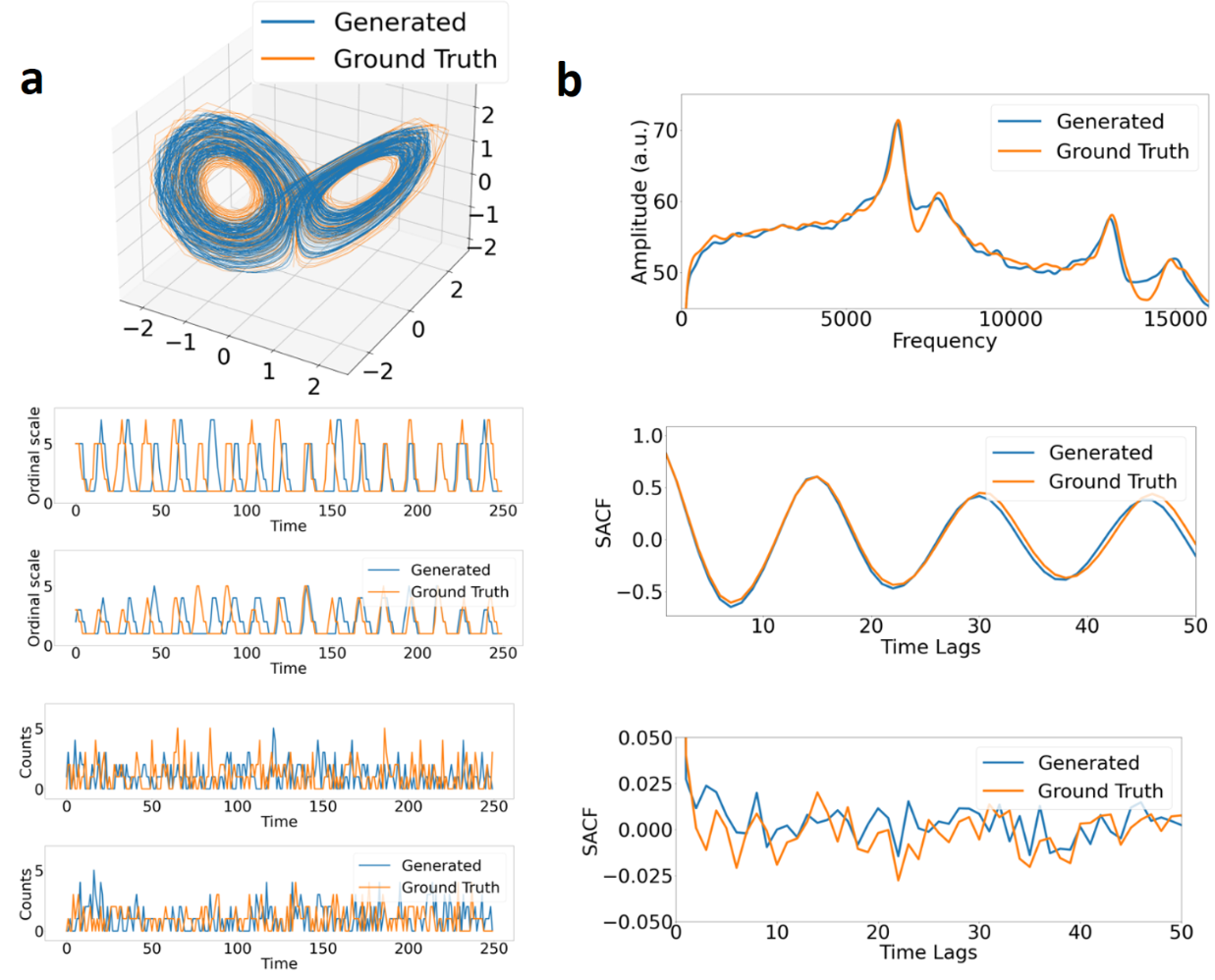}
	\caption{\textbf{a}: Freely generated example trajectories and time series from a dendPLRNN ($M=20,\ B=10,\  K=15,\ \tau=10$) trained with \ourmethodname\ jointly on Gaussian, ordinal, and count data sampled from a Lorenz-63 system. \textbf{b}: Example power spectra (Gaussian data) and Spearman autocorrelation functions (ordinal and count data). Simulated latent trajectories faithfully capture the geometry of the Lorenz attractor, as well as the temporal structure of the ground truth data when projected back into observation space.}
    \label{fig:example_trajectories}
\end{figure}

For this, we first sampled a time series of $100,000$ time steps and computed the dimension-wise Spearman autocorrelation for time lags up to $200$ for both generated and 
ground truth test data (see Fig. \ref{fig:example_trajectories}). We then calculate the average squared error between the corresponding SACFs of ground truth and generated time series across all lags and dimensions as our performance metric:
\begin{equation} 
\text{MSE}_{SACF} = \frac{1}{NT} \sum_{i=1}^{N} \sum_{\tau=1}^{T} \left( \text{SACF}_{\text{gen}, i}(\tau) - \text{SACF}_{\text{ground truth}, i}(\tau) \right)^2
\end{equation}

\paragraph{Mean squared prediction error} 
A mean squared prediction error (PE) was computed across test sets of length $T=10000$ by initializing the trained dendPLRNN with the test set time series up to some time point $t$ from the estimated initial state $\mathbf{E}[\mathbf{z}_t|\mathbf{x}_{1:t}]$, from where it was then iterated forward by $n$ time steps to yield a prediction at time step $t+n$. The $n$-step PE is then defined as the MSE between predicted and true observations:
\begin{equation}
PE(n)=\frac{1}{N(T-n)} \sum_{t=1}^{T-n}\sum_{i=1}^{N}(x_{i, t+n}-\hat{x}_{i, t+n})^2
\end{equation}
Due to exponential divergence of initially close trajectories in chaotic systems, the PE is sensible only for a limited number of time steps \citep{koppe_identifying_2019}; here we chose $n=10$.

 \paragraph{Determining initial states for computing mean squared prediction errors} 
   Given exponential trajectory divergence, the prediction error (PE) also depends on the precise initialization of the DS model at time $t$. 
   For SVAE training, an estimate of the initial state is directly provided by the encoder model. For BPTT-TF, the initial state $\mathbf{z}_t$ can be inferred by 
   a jointly trained linear mapping from $\mathbf{z}_t$ to $\mathbf{x}_t$ \citep{brenner22a}. For \ourmethodname\ and GVAE-TF, if $K=M$, the initial state $\mathbf{E}[\mathbf{z}_t|\mathbf{x}_{1:t}]$ is also directly given by the encoder model. If $K<M$, a subset of $M-K$ latent states is drawn from a standard normal distribution. In this under-specified case, we used a warm-up phase of $t_w$ time steps, where the system was initialized from the encoder at time $t-t_w$ and iterated forward, providing encoded states in the form of TF signals at every time step, which yields an initial state estimate $\mathbf{E}[\mathbf{z}_t|\mathbf{x}_{(t-t_w):t}]$. We determined $t_w=20$ by grid-search, leading to on average best results across datasets. We found that including this warm-up phase significantly improved prediction performance, although it still led to slightly worse results than for the fully specified case, $K=M$. Finally, for MS we neither have an encoder model nor a TF signal, and hence a sensible estimate for the initial state is difficult to obtain.

\paragraph{Ordinal prediction error} 
The ordinal PE was computed similarly as the mean squared PE above, but -- as pointed out in \citet{ogretir_variational_2022} -- due to the non-metric nature of ordinal data taking the absolute ($L_1$) deviation between observed and predicted values is more sensible:
\begin{equation}
OPE(n)=\frac{1}{N(T-n)} \sum_{t=1}^{T-n}\sum_{i=1}^{N}|o_{i, t+n}-\hat{o}_{i, t+n}|
\end{equation}

\paragraph{Spearman cross-correlation (SCC)}
To assess whether the global cross-correlation structure between the different ordinal time series is preserved by the reconstruction method, the Spearman correlation between each pair of ordinal time series was computed based on $100,000$ time steps long samples, using \texttt{scipy.stats.spearmanr}, for both generated and 
ground truth test data.
\begin{equation} \label{eq:SCC}
\text{SCC}_{ij} = \frac{1}{T} \sum_{t=1}^{T} \left(\frac{r_{it} - \overline{r}_i}{\sigma_{r_i}}\right) \left(\frac{r_{jt} - \overline{r}_j}{\sigma_{r_j}}\right)
\end{equation}
with $\sigma_{r_i}$ and $\sigma_{r_j}$ the standard deviation of the ranks of time series $i$ and $j$. We then calculated the MSE across all elements of the respective correlation matrices for the $N$ ordinal time series:
\begin{equation}
    \text{MSE}_{SCC} = \text{MSE}_{SCC} = \frac{1}{N^2-N} \sum_{i=1}^{N} \sum_{j\neq i}^{N} \left( \text{SCC}_{\text{gen}, ij} - \text{SCC}_{\text{ground truth}, ij} \right)^2
\end{equation}

 \paragraph{Lyapunov exponent}
The maximum Lyapunov exponent quantifies the divergence rate of nearby trajectories, and (for discrete-time systems $F_\vtheta\big(\vz_{t-1}, \vs_t \big)$ like the dendPLRNN, \Eqref{eq:plrnn_lat}) is defined as
\begin{align}\label{eq:lyap}
\lambda_{max} := \lim_{T \rightarrow \infty} \frac{1}{T} 
\log   \norm{ \ \prod_{r=0}^{T-2}  \mJ_{T-r} \ }_2,
\end{align}
where $\mJ_t := \frac{\partial\mF_\vtheta\big(\vz_{t-1}, \vs_t \big)}{\partial\vz_{t-1}} = \frac{\partial\vz_t}{\partial\vz_{t-1}}$ are the system's Jacobians and $\norm{\cdot}_2$ is the spectral norm.
To numerically approximate the Lyapunov spectrum for trained models, dendPLRNNs were iterated forward by $5500$ time steps using \Eqref{eq:plrnn_lat}, of which the first $500$ steps were discarded to remove transients. For numerical stability, an algorithm based on \citet{wolf_determining_1985, vogt_lyapunov_2022} was used which re-orthogonalizes the product series of Jacobians after every $5$-th time step using a QR decomposition. For consistency, we also computed maximum Lyapunov exponents for the Lorenz ($\lambda_{max}=0.903$), Rössler ($\lambda_{max}=0.071$), and Lewis-Glass ($\lambda_{max}=0.072$) models ourselves using the Julia library \texttt{DynamicalSystems.jl} \citep{Datseris2018} and the \texttt{dysts} Python package \cite{gilpin_chaos_2022}, which both use the same algorithm by \citet{wolf_determining_1985}, but our estimates agree closely with those that can be found in the literature (Lorenz: $\lambda_{max}=0.905$, Rössler: $\lambda_{max}=0.072$, in \citet{alligood_chaos_1996})

\subsection{Comparison methods}\label{sec:methods:comparisons}
As noted further above, different choices of DSR model are possible within our framework (see Appx. Table \ref{table_rnns}). Since the focus here is on evaluating the training framework for DSR from multimodal data itself, for comparability we used the same dendPLRNN model in all five other methods compared to. The total number of trainable parameters was kept approximately the same for all methods (see also Table \ref{tab:hypers}).

\paragraph{Multimodal SVAE}
First, we compared our method to a multimodal SVAE as proposed in \citep{kramer22a}, optimizing its multimodal ELBO. This, to our knowledge, is currently the only other general approach specifically designed for DSR from arbitrary data modalities observed simultaneously. We followed the implementation of the encoder model as provided on \url{https://github.com/DurstewitzLab/mmPLRNN}, training with a sequence length of $150$ time steps per batch, a hidden dimension of $20$, and the parameters of the dendPLRNN the same as for the runs with \ourmethodname.

\paragraph{Classical RNN training with truncated BPTT}
Second, we tried `classical' RNN training via truncated BPTT, where observations are provided as inputs at every time step, and the whole model is optimized using modality-specific decoder models, \Eqref{loss_PLRNN}. Specifically, given observed multimodal time series $\bm{Y}$ of length $T$, observations $\bm{y}_t$ are included into the RNN updates according to
\begin{align}\label{eq:plrnn_lat_input}
	\bm{z}_t = \bm{A} \bm{z}_{t-1} + \bm{W}  \phi(\bm{z}_{t-1}) + \bm{U y}_t +\bm{h}+ \bm{\epsilon}_t, \quad \bm{\epsilon}_t \sim\mathcal{N}(\bm{0},\bm{\Sigma})
	\end{align}
where $\bm{U}$ is the input-to-hidden weight matrix. Training is performed on short subsequences of length $T_{seq}$, where $T_{seq}$ is a hyper-parameter. The network then produces observations at each time step using the modality-specific decoder models, just as in MTF. During inference, the network is iterated forward by using the predicted observations at time $t$ as input for the predictions at time $t+1$. We probed different values for the sequence length $T_{seq} \in \{5 \dots 50\}$. We found best settings to agree with the TF interval $\tau$ for \ourmethodname\ and the sequence length $L$ for multiple shooting, but -- consistent with previous work \citep{mikhaeil_difficulty_2022} -- plain error gradient/ sequence truncation often led to divergence and failed to capture the true time scales in the fMRI data. 

\paragraph{Multiple shooting}
 A third approach that can deal with multimodal observation models but, unlike TF, does not require model inversion, is `multiple shooting (MS)' \citep{bock_multiple_1984}, a method suggested in the dynamical systems literature \citep{voss_nonlinear_2004}.
 Multiple shooting aims to solve boundary value problems by dividing time into sub-intervals, treating each sub-interval as a separate initial value problem, and then imposing continuity conditions between intervals. Multiple shooting has also been applied to DS reconstruction and Neural-ODE training \citep{iakovlev_latent_2022}, where the initial conditions (`shooting nodes') for each interval are model parameters and continuity across intervals is enforced through a penalty term in the loss \citep{voss_nonlinear_2004}. Hence, rather than controlling trajectory flows through a sparse TF signal applied after forcing intervals $\tau$, alternatively one may reset the latent model trajectory to an inferred initial condition after $\tau$ time steps. The advantage is that this method does not require the inversion of decoder models and is hence naturally suited to handle different data modalities without further care (i.e., retaining the distributional properties of the original data). 
More specifically, the observed time series $\bm{Y}$ is partitioned into $N_{seq}$ subsequences $\bm{Y}^s, s=1 \dots N_{seq}$, of length $L$, and for each subsequence a new initial condition $\bm{\mu}_0^s$ is learned. During training, trajectories are freely generated for $L$ time steps from $\bm{\mu}_0^s$ for each subsequence, and likelihoods for the observed trajectories $\bm{Y}^s$ are computed using the observation models from Eqs. \ref{eq:gauss_obs_model}-\ref{eq:poisson_obs_model}. A consistency (penalty) term in the loss ensures continuity between subsequences according to
\begin{equation} \label{eq:ms-loss}
\mathcal{L}_\text{MS}=\lambda_\text{MS} \sum_{s=1}^{N_{seq}-1}||F_\theta(\bm{z}_L^s)-\bm{\mu}_0^{s+1}||_2^2
\end{equation}
where $F_\theta$ in our case is the dendPLRNN, \Eqref{eq:plrnn_lat}, $\lambda_\text{MS}$ is a regularization parameter, and
$F_\theta(\bm{z}^{s}_L)=F_\theta(F_\theta(\dots F_\theta(\bm{\mu}_0^{s})))=F_\theta^L(\bm{\mu}_0^{s})$. The sequence length $L$ plays a similar role as the teacher forcing interval $\tau$ for MTF, controlling the times at which states and gradients are reset during training. Indeed, 
best performance is achieved when using choosing $L = \tau$ for all datasets studied here (see Table \ref{tab:hypers}).
    
\paragraph{Standard unimodal approach with data `Gaussianization'}\label{sec:methods:gaussian}
A naive approach for handling multimodal observations with any type of DSR model would be to pre-process all modalities such as to bring them into approximate agreement with Gaussian assumptions. Thus, for training the dendPLRNN with standard BPTT-TF \citep{brenner22a} and GVAE-TF, we transformed ordinal and count observations into approximately Gaussian variables through a Box-Cox-transformation \citep{box_analysis_1964}, $z$-scoring, and Gaussian kernel smoothing across the time series. For the optimal width of the Gaussian kernel, we performed a grid search over kernel sizes $\nu \in \{0, 0.01,0.1,1,10, 15, 20, 25\}$. Optimal settings for the results reported 
are given in Table \ref{tab:hypers}.

\subsection{Alternative encoder models}\label{sec:appx:encoders}

\paragraph{Mixture-of-Experts CNN encoder} \label{appx:mixture_of_experts}
The Mixture-of-Experts (MoE) approach \citep{shi_variational_2019} employs the same architecture as the CNN encoder described in Methods \ref{sec:methods:mvae}, but uses a distinct encoder for each modality. We combined the outputs of each encoder into a joint estimate using a weighted average: 
\begin{align*}
\bm{\mu}_{\text{MoE}} &= w_g\bm{\mu}_g + w_o\bm{\mu}_o + w_c\bm{\mu}_c \\
\bm{\Sigma}_{\text{MoE}} &= w_g\bm{\Sigma}_g + w_o\bm{\Sigma}_o + w_c\bm{\Sigma}_c,
\end{align*}
with means $\bm{\mu}_{\{g,o,c\}}$, diagonal covariances $\bm{\Sigma}_{\{g,o,c\}}$ and weights $w_g$ (Gaussian), $w_o$ (ordinal) and $w_c$ (count process), all set to $1/3$. 
We also tested a product-of-experts (PoE) approach \citep{hinton2002training, wu_multimodal_2018}, where the estimates of the individual experts are multiplied instead of summed, which however often led to numerical instabilities during training on discrete variables.

\paragraph{RNN encoder}
For the RNN encoder \citep{cho2014learning}, the hidden states $\bm{h}_t$ of an RNN, where observations $\bm{y}_t$ are provided as inputs at every time step (similar to \Eqref{eq:plrnn_lat_input}), are used to map onto the parameters of the approximate posterior at every time step. Here we use the standard RNN implementation in \texttt{torch.nn.RNN}, where
\begin{align*}
    \bm{h}_t &= \tanh(\bm{W}\bm{h}_{t-1} + \bm{U}\bm{y}_t + \bm{b}) \\
    \bm{\mu}_t &= \bm{W}_{\mu}\bm{h}_t + \bm{b}_{\mu} \\
    diag\left([\log \sigma_1^2, \ \dots, \ \log \sigma_{K}^2]\right) &= \bm{W}_{\Sigma}\bm{h}_t + \bm{b}_{\Sigma}
\end{align*}
with RNN parameters $\{\bm{W}, \bm{U}, \bm{b}\}$ and linear readout weights $\{\bm{W}_\mu, \bm{b}_\mu\}$ for the mean and $\{\bm{W}_\Sigma, \bm{b}_\Sigma\}$ for the logarithm of the diagonal covariance of the approximate posterior, respectively.

\paragraph{Transformer}
The Transformer encoder is based on the architecture in \cite{vaswani_attention_2017}. Since computation time scales with $T_{seq}^2$ in this architecture, we restricted the sequence length to $100$ steps. 
We used positional encodings to acknowledge the sequential (time series) nature of the data, as proposed in \cite{vaswani_attention_2017}. The sequence plus encodings is then passed through a standard Transformer encoder block using \texttt{torch.nn.TransformerEncoder}. As for the other encoder models, the output of the Transformer was mapped via linear readout layers onto the mean and logarithm of the covariance of the approximate posterior.

\paragraph{Multi-layer-perceptron (MLP)}
We also tested a standard MLP with $3$ layers and rectified linear unit (ReLU) nonlinearity as an encoder model, with outputs mapped via a linear readout layer to the mean and logarithm of the covariance of the approximate posterior, as for the other encoders tested.\\

Performance comparisons of these different encoder architectures are provided in Table \ref{table_encoders}.

\subsection{Datasets}\label{sec:methods:datasets}

\paragraph{Lorenz-63 system}\label{sec:supp:lorenz}
A stochastic version of the $3$d Lorenz-63 system, originally proposed in \citet{lorenz_deterministic_1963}, is defined by

 \begin{align} \label{eq:lorenz} \nonumber
       \mathrm{d}x &= (\sigma (y - x))\mathrm{d}t+\mathrm{d}\epsilon_1(t), \, \\
        	\mathrm{d}y &= (x (\rho - z) - y) \mathrm{d}t+\mathrm{d}\epsilon_2(t), \, \\\nonumber
       \mathrm{d}z &=  (x y - \beta z) \mathrm{d}t+\mathrm{d}\epsilon_3(t). \\ \nonumber
    \end{align}

Parameters used for producing ground truth data in the chaotic regime were $\sigma=10, \rho=28$, and $\beta=8/3$. Process noise was injected into the system by drawing from a Gaussian $\mathrm{d}\boldsymbol\epsilon \sim\mathcal{N}(\bm{0},0.01^2 \mathrm{d}t \times \bm{I})$. For both training and test data, a trajectory of $100,000$ time steps was sampled, performing numerical integration with \texttt{scipy.odeint} ($\mathrm{d}t=0.05$; note this value differs from the one used in \citet{mikhaeil_difficulty_2022}, explaining why different $\tau$ values were required for optimal reconstructions). To obtain multimodal observations, trajectories drawn from the ground truth system were fed into the different types of observation models in Eqs. \ref{eq:gauss_obs_model}-\ref{eq:poisson_obs_model}, with randomly drawn parameters.

\paragraph{Symbolic representation of Lorenz-63 dynamics} \label{sec:supp:symbolic}
To obtain a symbolic representation of the Lorenz-63 system, we first sampled a time series as above, using \Eqref{eq:lorenz}, and removed transients. We then divided the system's state space into $N^3$ cubes (with $N=4$ bins per dimension). 
The range on each axis covered by the bins was set to the minimum-to-maximum extent of the attractor. Each data point of the time series was then assigned a label corresponding to the cube it fell in, using a one-hot encoding of length $N^3$. In fact, as for the Lorenz-63 chaotic attractor $28$ of the $64$ bins turned out empty (did not contain any points of the time series), the number of symbols (i.e., vector length) could be reduced further to just $36$ (see also Fig. \ref{fig:discrete-DSR} for decoded bin probabilities). The procedure is sketched in the bottom half of Fig. \ref{fig:discrete-DSR}. We then furthermore performed a delay-embedding of the symbolic time series using a delay $\tau=10$ and an embedding dimension of $d=2$, thus doubling the length of the symbolic vector ($d=3$ gave similar results; see \citet{matilla-garcia_selection_2021} for a discussion of how to determine optimal delay embedding parameters for symbolic sequences, although not used for the present purposes). Although not strictly necessary, this step profoundly improved DS reconstruction from symbolic sequences.

\paragraph{Rössler system}\label{sec:supp:roessler}
The Rössler system was introduced 
in \citet{rossler_equation_1976} as a simplified version of the Lorenz system, and is given (in SDE form) by

 \begin{align} \label{eq:roessler} \nonumber
       \mathrm{d}x &= (-y - z)\mathrm{d}t+\mathrm{d}\epsilon_1(t), \, \\
        	\mathrm{d}y &= (x + ay) \mathrm{d}t+\mathrm{d}\epsilon_2(t), \, \\\nonumber
       \mathrm{d}z &=  (b + z(x-c)) \mathrm{d}t+\mathrm{d}\epsilon_3(t). \\ \nonumber
    \end{align}

Parameters used for producing ground truth data in the chaotic regime were $a=0.2, b=0.2$, and $c=5.7$. Process noise was added by drawing $\mathrm{d}\boldsymbol\epsilon \sim\mathcal{N}(\bm{0},0.01^2 \mathrm{d}t \times \bm{I})$. Training and test data was sampled as described above for the Lorenz-63 system, using $\mathrm{d}t=0.1$.

\paragraph{Lewis-Glass chaotic network model}\label{sec:supp:hopfield}

\begin{figure}[!htb]
\centering
	\includegraphics[width=0.6\linewidth]{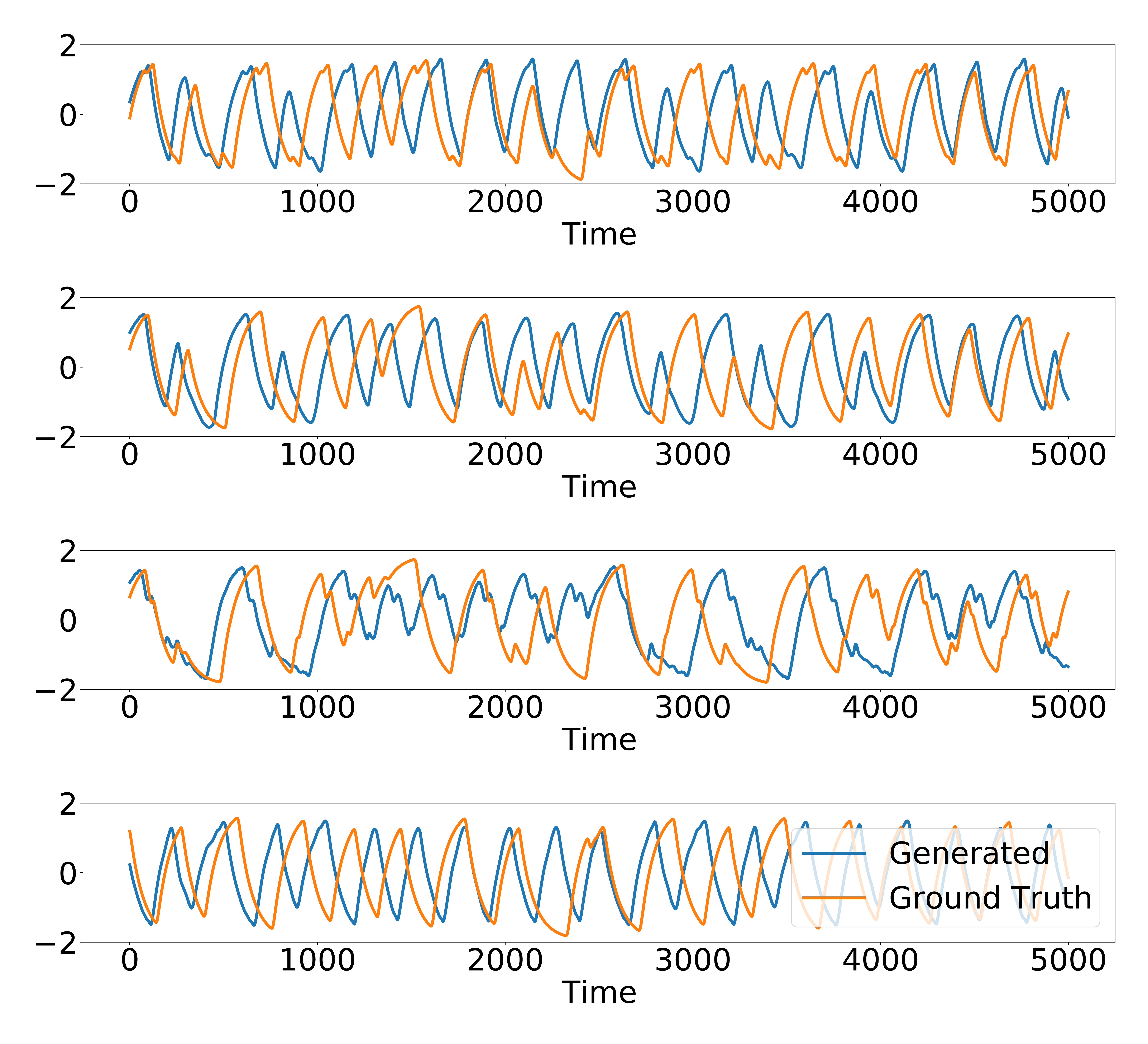}
	\caption{Example ground truth time series and freely generated series from a dendPLRNN ($M=20, B=10, K=15, \tau=20$) trained with \ourmethodname\ on the Lewis-Glass neural network model \citep{lewis_nonlinear_1992}.}
    \label{fig:hopfield_reconstructions}
    \end{figure}

We simulate a $6$-dimensional model of a neural network, originally introduced in \citet{lewis_nonlinear_1992}. Here the individual units of the network are endowed with a continuous gain function $G(x)=\frac{1+\tanh(-\alpha x)}{2}$, with the vector field given by
\begin{equation}
    \frac{\mathrm{d}\bm{x}}{\mathrm{d}t}=\frac{-\bm{x}}{\tau} + 
    G(\epsilon \, \bm{K} \bm{x}) - \beta .
\end{equation}
To sample from this system in the chaotic regime, we used the \texttt{Hopfield} model implementation in the Python package \texttt{dysts.flows}, based on \citet{gilpin_chaos_2022}. Here, $\alpha=-1, \beta=0.5, \epsilon=10, \tau=2.5$, and
\begin{align*}
\bm{K}=\begin{bmatrix}
0 & -1 & 0 & 0 & -1 & -1 &\\
0 & 0 & 0 & -1 & -1 & -1 &\\
-1 & -1 & 0 & 0 & -1 & 0 &\\
-1 & -1 & -1 & 0 & 0 & 0 &\\
-1 & -1 & 0 & -1 & 0 & 0 &\\
0 & -1 & -1 & -1 & 0 & 0 &\\
\end{bmatrix}
\end{align*}
We generated training and test data by \texttt{maketrajectory}, and down-sampled the generated data by a factor of $30$. We sampled ordinal and count data in the same way as for the other datasets. Example time series and reconstructions are displayed in Fig. \ref{fig:hopfield_reconstructions}.

\begin{figure*}[!htb]
\centering
	\includegraphics[width=0.99\linewidth]{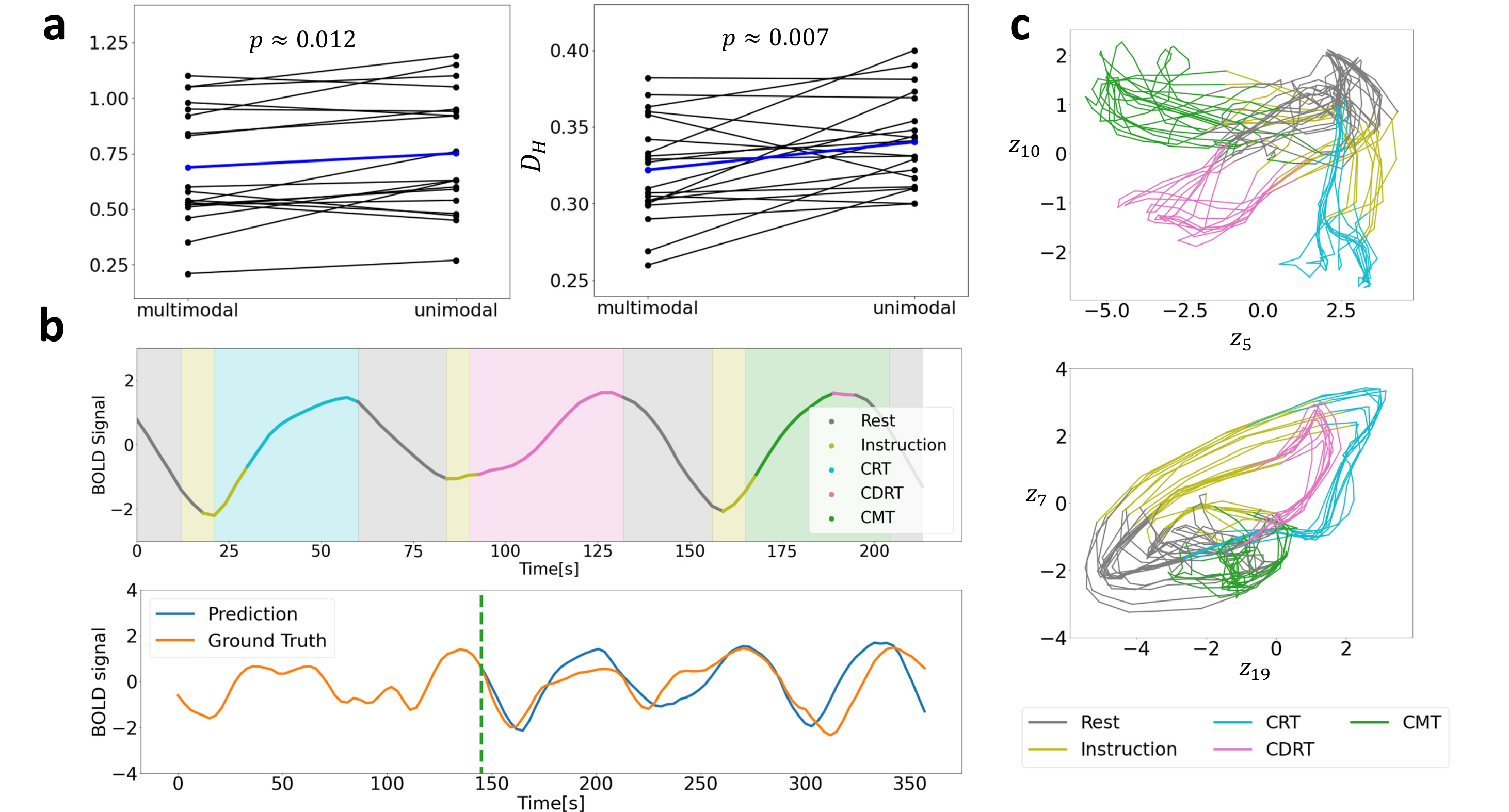}
	\caption{\textbf{a}: Multimodal integration using \ourmethodname\ on functional magnetic resonance imaging (fMRI)+behavioral data significantly helps to improve DS reconstruction compared to just fMRI data alone (unimodal). Results are for $20$ subjects (black lines; blue line = mean), showing geometrical ($D_{stsp}$, left) and temporal ($D_H$, right) disagreement between true and reconstructed systems; $p$-values from paired t-tests as indicated. 
 \textbf{b}: Example of decoded (color-coding of time series) and true (pale background colors) task stages for one brain area and subject. The trained model was freely iterated forward from the first time step of the test set, and task stages were decoded from the simulated activity. Decoding of class label $\hat{l} \in $\{Rest, Instruction, CRT, CDRT, CMT\} is based on the maximum posterior probability, 
 $\hat{l}_t={\arg\max} \ p(l_{kt}|\bm{z}_t)$, given the latent trajectory $\bm{z}_t$. The graph below shows true and predicted BOLD signals. 
 \textbf{c}: 2d subspaces of freely generated latent activity for a dendPLRNN ($M=30, B=10, K=20, \tau=7$) trained jointly on continuous and categorical data by \ourmethodname\ for one example subject. The color coding corresponds to the task labels predicted according to the maximum posterior probability given the latent state as in \textbf{b} 
 at each time step. The latent space appears to be structured according to the different task stages.}
 \label{fig:fmri_main}
    \end{figure*}

\paragraph{Human fMRI dataset} \label{sec:supp:fmri}
This same dataset was used previously in the study by \citet{kramer22a} on multimodal data integration for DS reconstruction, and is openly available at \url{https://github.com/DurstewitzLab/mmPLRNN}. 
In the study, $26$ participants (of which $20$ were selected for analysis, cf. \citet{kramer22a}) were shown a series of images of different shapes (rectangles and triangles) while undergoing fMRI 
with a sampling rate of $1/3$ \si{\hertz}. The subjects were asked to identify the type of shape presented in the current image or in the preceding one, respectively, under three different task conditions: a continuous delayed response 1-back task (CDRT), a continuous matching 1-back task (CMT), and a 0-back control choice reaction task (CRT). A resting condition and an instruction phase were also included. Each task condition was repeated five times, where the last repetition of all task stages was left out as a test set in our analyses (see Fig. \ref{fig:fmri_main}\textbf{b}). The brain activity of the participants as assessed via the BOLD 
signal projected onto the first principal component within each of $10$ different brain regions in both hemispheres that were identified to be relevant for this task. More details on this study can be found in \citet{koppe_temporal_2014}.
Due to the short length of these time series, 
for the fMRI results from Table \ref{table_fmri} we chose a sequence length of $72$ time steps for testing and training.

\paragraph{Hippocampal multiple single-unit (MSU) and spatial position data}\label{sec:methods:hippo}
For our second empirical example, we used electrophysiological recordings from the rodent hippocampus and spatial location data \citep{grosmark_dataset_2016}, publicly available at \url{https://crcns.org/data-sets/hc/hc-11/about-hc-11}. Specifically, the ``sessInfo.mat'' file from rat `Achilles', which contains extracted spike times, was chosen for exemplification and further preprocessed using the script from \citep{zhou_learning_2020}, provided at \url{https://github.com/zhd96/pi-vae/blob/main/code/rat_preprocess_data.py}, to obtain counts per time interval, using a binning width of $200$ ms. For our analysis, we focused exclusively on the MAZE task 
and selected the $60$ most active neurons, as many neurons had very low (statistically insufficient) activity levels. The position data was represented as a continuous $1$d vector, with missing values imputed from neighboring points. 
Rats received water rewards at both end points of the track, which were provided to the model as scalar external signals $s_t$ (cf. \Eqref{eq:plrnn_lat}) 
set to $1$ for 5 time bins around the time points at which the rat started to move away from the reward location 
(with trainable weights $\bm{U} \in \mathbb{R}^{M \times 1}$ in \Eqref{eq:plrnn_lat}).

\section{Supplemental Results} \label{sec:supp:results}

\begin{figure*}[!htb]
\centering
	\includegraphics[width=0.99\linewidth]{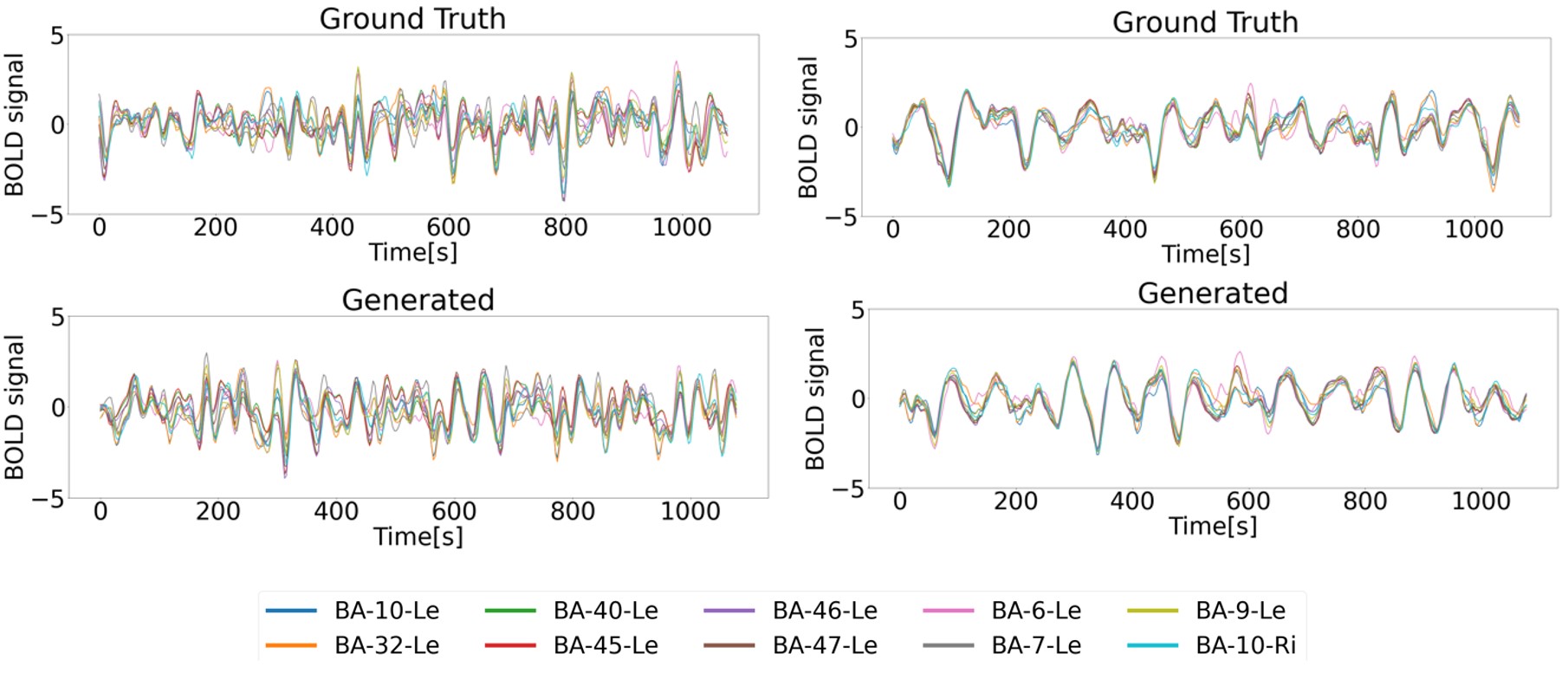}
	\caption{Freely generated time series from $10$ brain areas per subject from subjects $\#3$ (left) and $\#7$. (right). The trained DSR model, only iterated by providing an initial state, captures the overall temporal structure of the complex activity patterns even from very short time series.}
 \label{fig:fmri_freely_gen}
    \end{figure*}

\begin{figure}[!htb]
    
    \centering
	\includegraphics[width=0.8\linewidth]{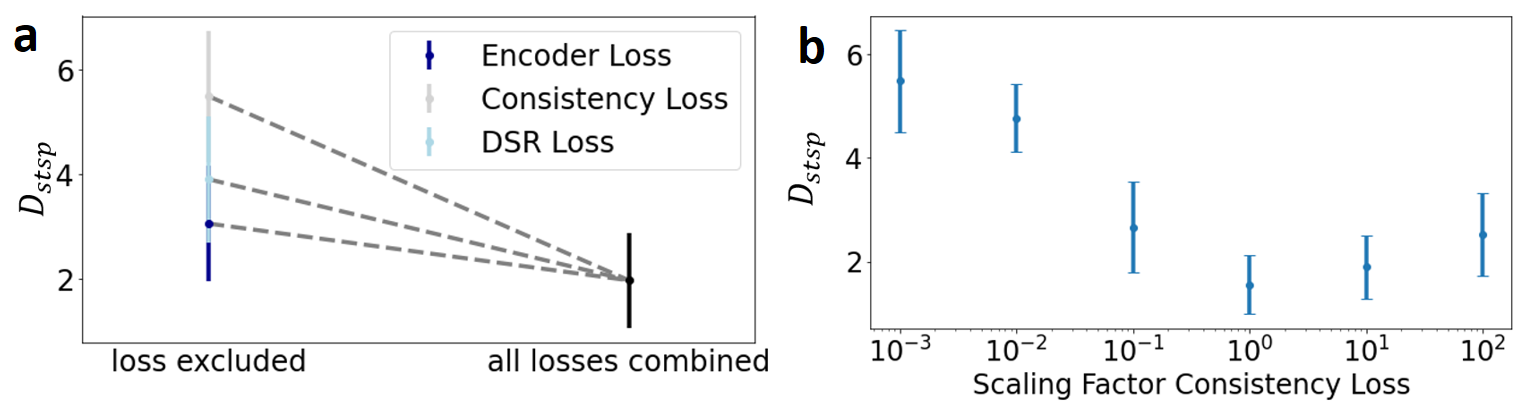}
 \caption{Impact of different loss terms in \Eqref{eq:loss_total} on DS reconstruction quality, indicating that all loss components are crucially important. \textbf{a}: Comparison of state space agreement $D_{stsp}$ 
 when excluding different terms from the total loss in \Eqref{eq:loss_total}, and \textbf{b} for different scaling of the consistency loss ($\mathcal{L}_\text{con}$), for the dendPLRNN trained by \ourmethodname\ on data from the chaotic Lorenz system as in Sect. \ref{sec:exp:benchmarks}. 
 }
\label{fig:loss_scaling}
\end{figure}

\begin{figure*}[!htb]
    \centering
	\includegraphics[width=0.8\linewidth]{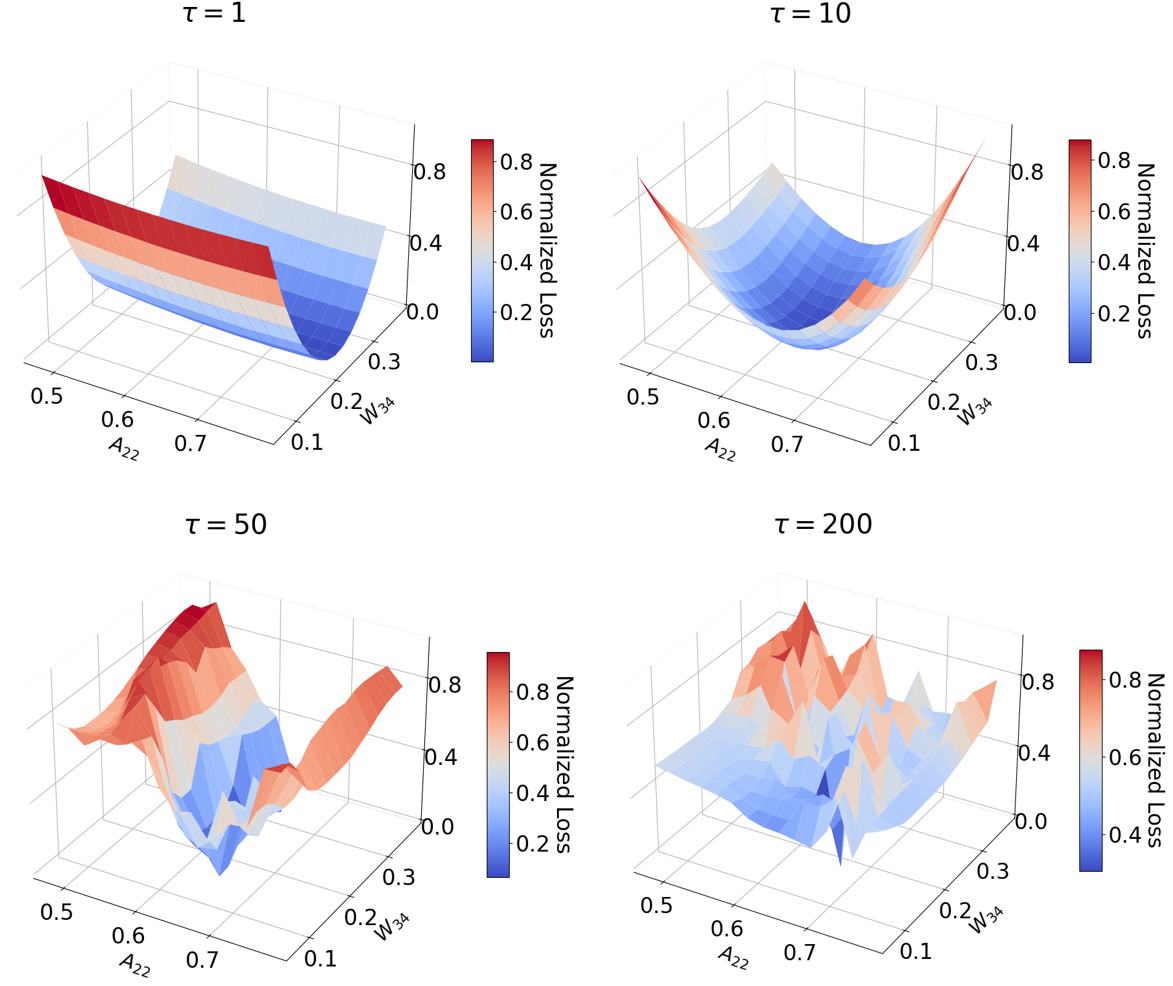}
	\caption{\ourmethodname\ loss landscapes (\Eqref{eq:loss_total}) as a function of two randomly chosen parameters ($A_{22}$, $W_{34}$) of the DSR model (\Eqref{eq:plrnn_lat}) for four different settings of the TF interval $\tau$. The loss was computed on a random batch of training data for the model from Fig. \ref{fig:noisy_lorenz}. Smoother loss landscapes are obtained for lower $\tau$ (similar as observed for GTF in \citet{pmlr-v202-hess23a}). While for the optimal TF interval $\tau=10$ the loss is smooth and convex, for too low $\tau=1$ flat directions in the loss appear.
 }
	\label{fig:losslandscape}
\end{figure*}

\begin{figure}[!htb]
  \centering \includegraphics[width=1.0\textwidth]{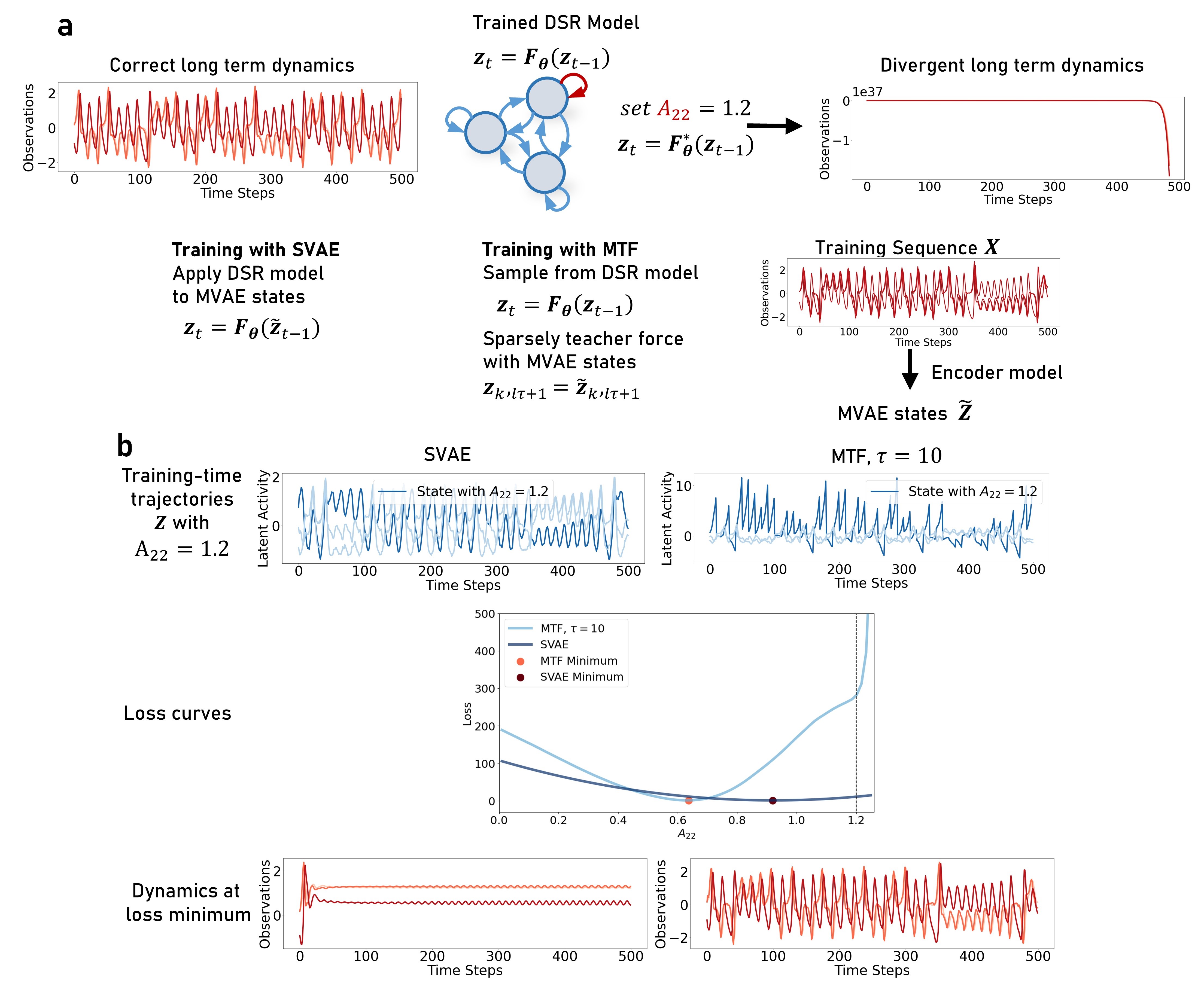}
  \caption{Illustration of how dramatic changes in long-term dynamics differently impact the SVAE and MTF loss. 
  \textbf{a}: A dendPLRNN successfully trained on multimodal observations from the Lorenz-63 system (Fig. \ref{fig:noisy_lorenz}) is altered by setting a parameter of the linear self-connectivity $A_{22}>1$, which results in globally diverging dynamics, yet locally still consistent with the Lorenz-63.
  \textbf{b}: 
  The global divergence is clearly apparent in the 
  MTF training-time trajectories $\bm{Z}$ generated using STF with interval $\tau=10$ (right), within which the DSR model evolves freely, unlike the SVAE (left). 
  This divergence leads to large increases in the MTF training loss (see MTF loss curve for $A_{22}>1$), and hence is strongly penalized. 
  This effect is essentially not present for the SVAE, where the global divergence induces no considerable effect on the training loss. 
  Hence, the mismatch in global (long-term) dynamics remains unrecognized by the SVAE, explaining why the MTF approach is so superior. 
  In fact, as shown at the bottom, at the minimum of the SVAE loss ($A_{22} \approx 0.966$), only short-term predictions are correctly reproduced (left), while the MTF at its minimum ($A_{22} \approx 0.637$) produces trajectories whose long-term dynamics agree in their temporal structure with those of the original Lorenz-63 (right).
  } \label{svae_mtf_comparison}
\end{figure}

\begin{figure}[!htb]
\centering
	\includegraphics[width=0.65\linewidth]{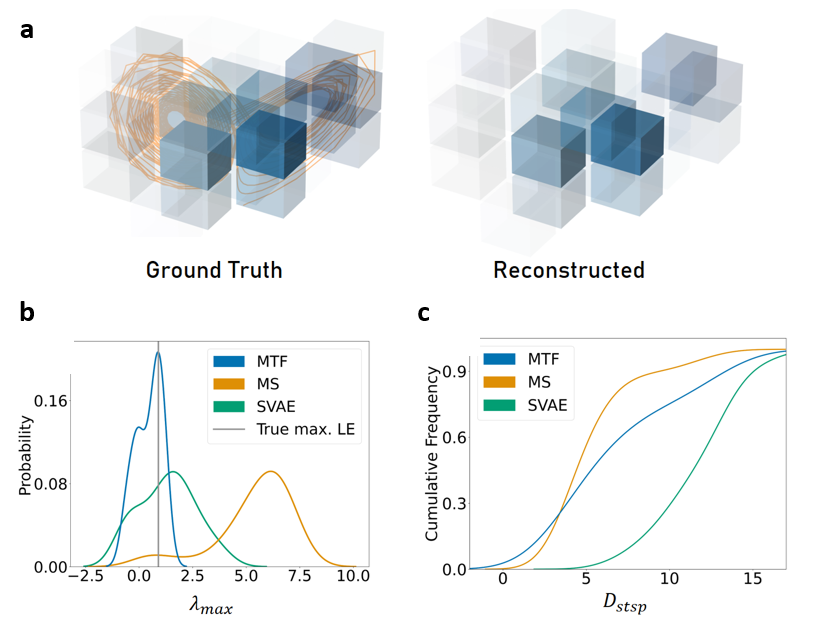}
	\caption{\textbf{a}: True and predicted class label probabilities (given the maximum posterior probability for a category at each time step) from a freely generated trajectory of a dendPLRNN ($M=30, B=15, K=30, \tau=10$), trained with MTF on the symbolic representation of the chaotic Lorenz-63 dynamics.
 \textbf{b} and \textbf{c}: Kernel-density estimates of distributions of maximum Lyapunov exponents (\textbf{b}) and cumulative distributions of $D_{stsp}$ (\textbf{c}), comparing training with MTF, Multiple Shooting (MS), and multimodal SVAE across $30$ trained models, all using the same dendPLRNN model. While for the \ourmethodname\ the estimates of Lyapunov exponents are centered around the true value ($\lambda_{max}=0.903$), for SVAE and MS the distributions were much farther off.}
 \label{fig:symbolic_lorenz_categories}
\end{figure}

\begin{figure}[!htb]
\centering
	\includegraphics[width=0.99\linewidth]{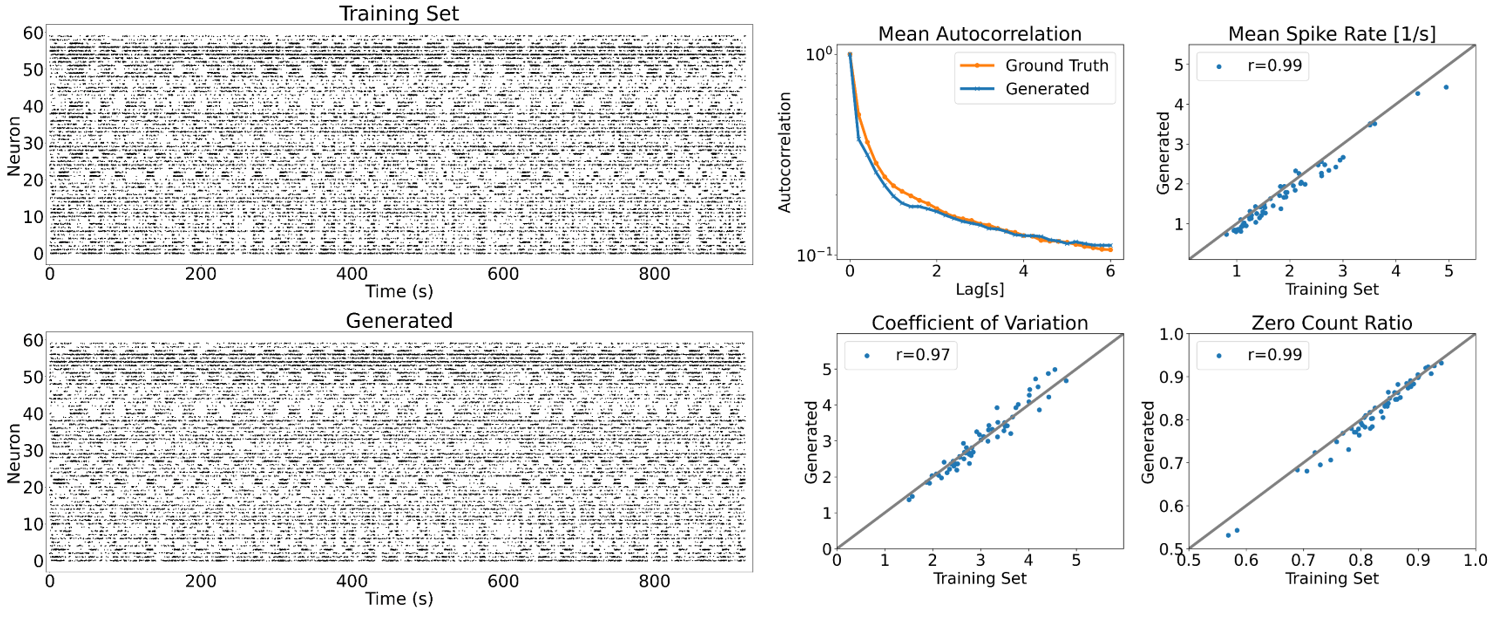}
	\caption{Example reconstructions of spike trains and spike statistics on the training set (see Methods \ref{sec:methods:hippo}). Test set reconstructions and further statistics are in Figure \ref{fig:hippocampus_main}.}
    \label{fig:hippo_train}
    \end{figure}

\begin{table}[!htb]
\caption{Hyperparameter settings for \ourmethodname, SVAE, BPTT, GVAE-TF, and MS trained on the Lorenz, Rössler and Lewis-Glass model, and for the experimental fMRI data. For explanation of symbols, see Methods \ref{sec:methods:comparisons}.}
\centering
\begin{tabular}{ c c c c c c c}
Dataset & M & B & K &$\tau,L, T_{seq}$ & $\lambda_\text{MS}$ &$\nu$ \\ 
\hline
Lorenz &  $20$ & $15$ & $15$ & $10$ & $1.0$ & $10$ \\ 
Rössler &  $20$ & $15$  & $15$ & $10$ & $1.0$ & $15$ \\ 
Lewis-Glass &  $20$ & $15$  & $15$ & $20$ & $1.0$ & $20$ \\ 
fMRI &  $30$ & $40$  & $30$ & $7$ & $1.0$ & $1$ \\ 
        \bottomrule
\end{tabular}
\label{tab:hypers}
\end{table}

\begin{table*}[!htp]
\caption{Performance comparison of different encoder models used in the MTF framework, trained on multimodal data from the chaotic Lorenz system. PE is the prediction error and OPE the ordinal prediction error. Note that SCC (Spearman cross-correlation), OACF (ordinal autocorrelation function), and CACF (count autocorrelation function) all refer to mean-squared-errors (MSEs) between ground truth and generated correlation functions. 
\\}
\centering
\scalebox{0.8}{
\begin{tabular}{lcccccccc}
\hline
Encoder  & $D_{stsp}$ $\downarrow$ & $D_{H}$ $\downarrow$ & PE $\downarrow$ & OPE $ \downarrow$ & SCC $\downarrow$ & OACF $\downarrow$  & CACF $\downarrow$ \\ \hline
CNN & $\mathbf{3.4 \pm 0.35}$   & $\mathbf{0.30 \pm 0.06}$  & $\mathbf{1.3\mbox{e$-$}2} \pm \mathbf{2\mbox{e$-$}4}$  & $\mathbf{0.12 \pm 0.03}$  &  $\mathbf{0.07 \pm 0.01}$  &  $\mathbf{0.07 \pm 0.01}$   &  $\mathbf{6.6\mbox{e$-$}5 \pm 8\mbox{e$-$}6}$  \\
CNN-MoE & $5.89 \pm 0.18$ & $0.43 \pm 0.03$ & ${2.3\mbox{e$-$}2} \pm {5\mbox{e$-$}4}$  & \textbf{$0.13 \pm 0.00$} & $0.10 \pm 0.00$ & $0.19 \pm 0.01$ & ${1.1\mbox{e$-$}4 \pm 2\mbox{e$-$}5}$ \\
RNN & $5.47 \pm 0.48$ & $0.32 \pm 0.04$ & ${1.6\mbox{e$-$}2} \pm {2\mbox{e$-$}4}$ & $0.15 \pm 0.01$ & $0.13 \pm 0.02$ & $0.05 \pm 0.01$ & ${8.5\mbox{e$-$}5 \pm 9\mbox{e$-$}6}$ \\
Transformer & $5.85 \pm 0.14$ & $0.40 \pm 0.04$ & ${4.8\mbox{e$-$}2} \pm {5\mbox{e$-$}4}$ & $0.16 \pm 0.00$ & $0.17 \pm 0.03$ & $0.16 \pm 0.02$ & ${9.5\mbox{e$-$}5 \pm 7\mbox{e$-$}6}$ \\
MLP & $6.57 \pm 0.14$ & $0.43 \pm 0.01$ & ${5.4\mbox{e$-$}2} \pm {6\mbox{e$-$}4}$ & $0.15 \pm 0.00$ & $0.15 \pm 0.01$ & $0.21 \pm 0.01$ & ${1.3\mbox{e$-$}4 \pm 9\mbox{e$-$}6}$ \\
\hline
\end{tabular}
}
\label{table_encoders}
\end{table*}

\begin{table*}[!htb]
\caption{Performance comparison of different RNNs used as DSR model in the MTF framework, incl. the dendPLRNN \citep{brenner22a} used here, an LSTM \citep{hochreiter_lstm_97}, and a GRU \citep{chung_empirical_2014}, 
all trained by MTF on multimodal data from the chaotic Lorenz system ($10 \%$ Gaussian observation noise). PE is the prediction error and OPE the ordinal prediction error. Note that SCC (Spearman cross-correlation), OACF (ordinal autocorrelation function), and CACF (count autocorrelation function) all refer to mean-squared-errors (MSEs) between ground truth and generated correlation functions. \\
}
\centering
\scalebox{0.8}{
\begin{tabular}{lcccccccc}
\hline
RNN Model  & $D_{stsp}$ $\downarrow$ & $D_{H}$ $\downarrow$ & PE $\downarrow$ & OPE $ \downarrow$ & SCC $\downarrow$ & OACF $\downarrow$  & CACF $\downarrow$ \\ \hline
dendPLRNN & $\mathbf{3.4 \pm 0.35}$   & $\mathbf{0.30 \pm 0.06}$  & $\mathbf{1.3\mbox{e$-$}2 \pm 2\mbox{e$-$}4}$  & $\mathbf{0.12 \pm 0.03}$  &  $\mathbf{0.07 \pm 0.01}$  &  $\mathbf{0.07 \pm 0.01}$   &  $\mathbf{6.6\mbox{e$-$}5 \pm 8\mbox{e$-$}6}$  \\
LSTM & \textbf{$3.8 \pm 0.74$} & $0.31 \pm 0.01$ &  ${5.4\mbox{e$-$}2} \pm {5\mbox{e$-$}4}$ & ${0.16 \pm 0.03}$ & \textbf{$0.09 \pm 0.02$} & $0.09 \pm 0.02$ & ${8.8\mbox{e$-$}5 \pm 8\mbox{e$-$}6}$ \\
GRU & $\mathbf{3.47 \pm 0.56}$ & $\mathbf{0.29 \pm 0.03}$ & ${3.5\mbox{e$-$}2} \pm {5\mbox{e$-$}4}$ & ${0.13 \pm 0.03}$ & $\mathbf{0.06 \pm 0.01}$ & $\mathbf{0.08 \pm 0.01}$ & ${7.1\mbox{e$-$}5 \pm 5\mbox{e$-$}6}$ \\
\hline
\end{tabular}
}
\label{table_rnns}
\end{table*}

\begin{table*}[!htb]
\caption{Comparison of dendPLRNN, \eqref{eq:plrnn_lat}, trained by \ourmethodname\ (proposed method), by a sequential multimodal VAE (SVAE) based on \cite{kramer22a}, and a multiple-shooting (MS) approach, on $8$ ordinal observations with seven ordered categories, produced by the chaotic Lorenz system, Rössler system, and Lewis-Glass model, and on a symbolic representation of the chaotic Lorenz system. Values are mean $\pm$ SEM, averaged over 15 trained models. X = value cannot easily be computed for MS (because here initial conditions cannot be obtained directly from the data but require additional parameters). \\}
\centering
\renewcommand{\arraystretch}{1.2} 
\scalebox{0.7}{
\begin{tabular}{lc c c c c c }
\hline
Dataset& Method  & $D_{stsp}$ $\downarrow$ & $\lambda_{max}$ & OPE $\downarrow$ & SCC $\downarrow$ & OACF $\downarrow$ \\ \hline
\multirow{3}{*}{Lorenz-ordinal}  & \ourmethodname\ & $\mathbf{8.8 \pm 0.59}$  &$\mathbf{0.92 \pm 0.39}$ & $\mathbf{0.24 \pm 0.015}$  &  $\mathbf{0.085 \pm 0.02}$ &  $\mathbf{0.016 \pm 0.04}$ \\ 
                         & SVAE    & $14.7 \pm 0.7$  &$0.44 \pm 0.71$ & $0.8 \pm 0.03$   &  $0.17 \pm 0.02$  &  $0.23 \pm 0.02$ \\
                         & MS    & $13.8 \pm 1.1$   &$0.47 \pm 0.67$ &  X   &  $0.24 \pm 0.06$  &  $0.15 \pm 0.03$ \\\hline                      
\multirow{3}{*}{Rössler-ordinal} & \ourmethodname\ & $\mathbf{7.9 \pm 0.8}$ &$\mathbf{0.03 \pm 0.07}$ & $\mathbf{0.093 \pm 0.007}$  &  $\mathbf{0.051 \pm 0.009}$&  $\mathbf{0.051 \pm 0.009}$\\                     
                         & SVAE    & $11.5 \pm 1.3$ &-$0.27 \pm 0.58$ & $0.39 \pm 0.02$  &  $0.23 \pm 0.05$ &  $0.18 \pm 0.04$ \\ 
                        & MS    & $14.1 \pm 1.0$ &-$0.05 \pm 0.12$ & X  &  $0.12 \pm 0.04$ &  $0.14 \pm 0.03$ \\ \hline
\multirow{3}{*}{Lewis-Glass-ordinal} & \ourmethodname\ & $\mathbf{0.89 \pm 0.04}$  &$\mathbf{-0.11 \pm 0.41}$ & $\mathbf{0.15 \pm 0.02}$  &  $\mathbf{0.28 \pm 0.05}$&  $\mathbf{0.15 \pm 0.03}$\\ 
                         & SVAE    & $1.40 \pm 0.22$ &$ -1.8 \pm 2.1$ & $0.29 \pm 0.01$  &  $0.49 \pm 0.04$ &  $0.24 \pm 0.02$ \\
                         & MS    & $1.0 \pm 0.14$  &$-0.14 \pm 0.31$ & X  &  $0.51 \pm 0.04$ &  $0.45 \pm 0.03$ \\ \hline
\multirow{3}{*}{Lorenz-symbolic}  & \ourmethodname\ & $\mathbf{4.4 \pm 2.69}$   & $\mathbf{0.67 \pm 0.37}$ & & & \\ 
                         & SVAE    & $12.02 \pm 1.98$   & $1.87 \pm 0.88$  & & &  \\
                         & MS    & $\mathbf{4.46 \pm 1.82}$   &  $5.67 \pm 1.25$ & & &  \\\hline                       
\end{tabular}
}
\label{table_discrete}
\end{table*}

\begin{figure}[!htb]
    
    \centering
	\includegraphics[width=0.75\linewidth]{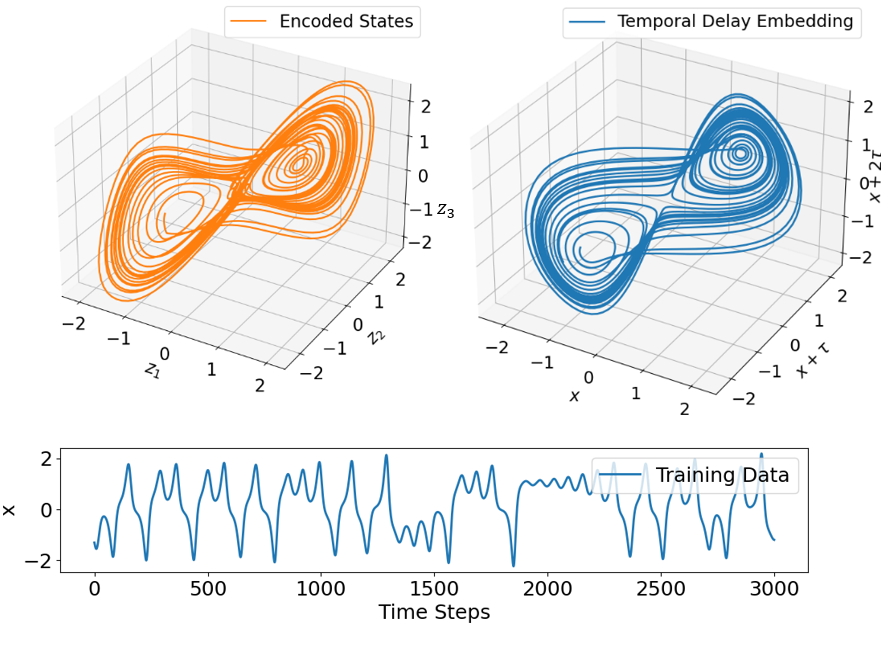}
 \caption{Reconstruction with missing variables (partial observations). Mean encoded states $\langle \tilde{\bm{Z}} \rangle$ 
 (left) after training a shPLRNN \cite{pmlr-v202-hess23a} with MTF on just one-dimensional time series from the Lorenz-63 system ($x$ coordinate, bottom). Note that the reconstructed attractor's geometry closely resembles that of the true Lorenz-63 obtained from an optimal temporal delay embedding of the $x$ variable (right; time lag chosen as the first minimum of the mutual information, see \citet{kantz_nonlinear_2004}). This illustrates that MTF can perform a kind of implicit delay embedding when the underlying system is not fully observed. 
 }
\label{fig:mtf_temp_convolutions_tde}
\end{figure}

\begin{figure}[!htb]
  \centering \includegraphics[width=1.0\textwidth]{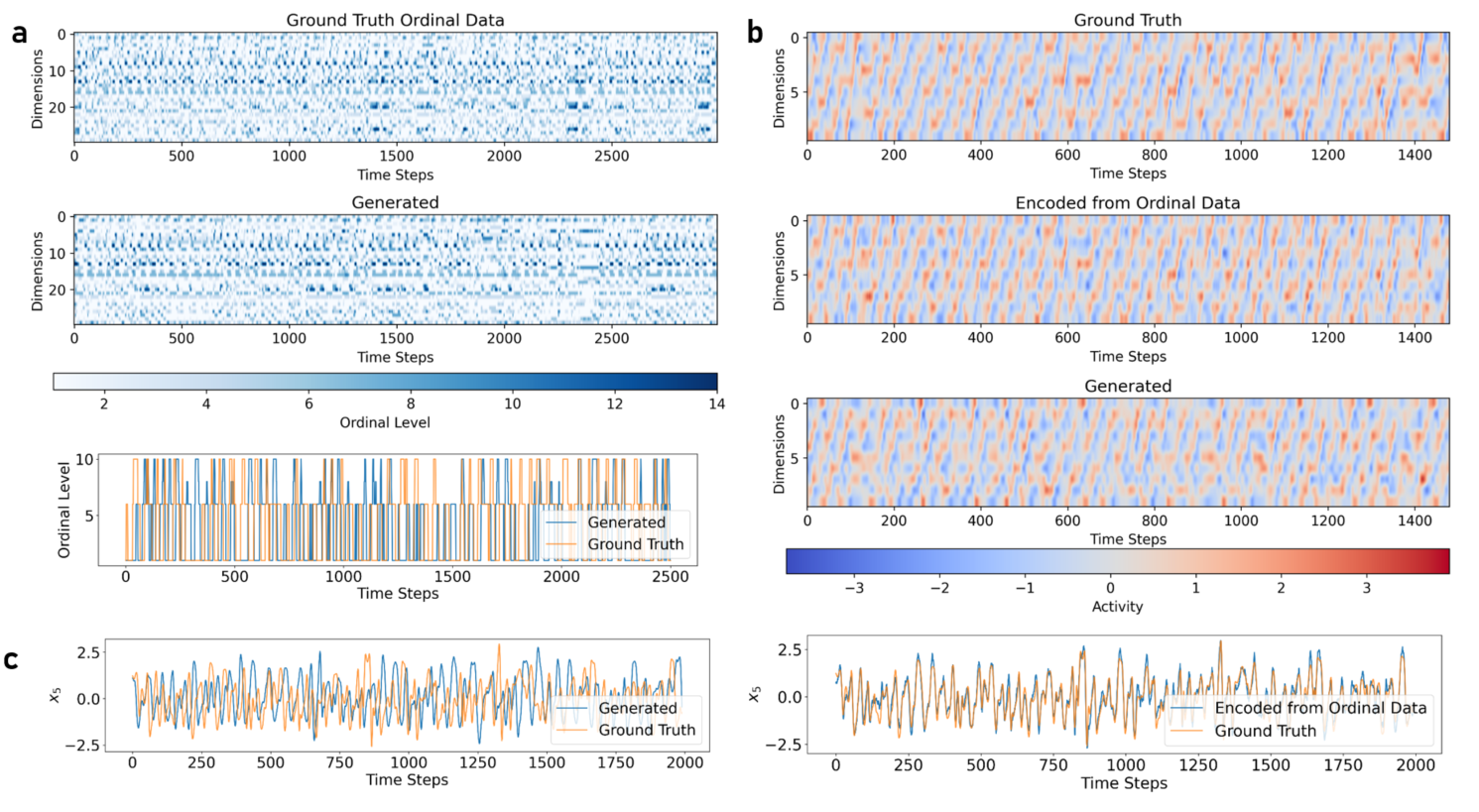}
  \caption{Reconstruction of a $10$d chaotic Lorenz-96 system solely from ordinal observations with up to $15$ levels using a shPLRNN (\citet{pmlr-v202-hess23a}; $M=10, L=100, \tau=10$) as DSR model. On average, each variable only occupied 7 unique categories due to the random initialization of the observation model. \textbf{a}, top: Ground truth ordinal time series sampled from a randomly initialized ordinal observation model $p({\bm{o}}_t|\bm{x}_t)$ from ground truth states $\bm{x}_t$ of the Lorenz-96 system, and reconstructed (generated) ordinal observations using the trained decoder model $p({\bm{o}}_t|\tilde{\bm{z}}_t)$. Bottom: Example ground truth and freely generated ordinal time series from 1 channel.
  \textbf{b}: Ground truth states $\bm{x}_t$ (top), states encoded using the mean of the states taken from the trained MTF encoder $p(\tilde{\bm{z}}_t|\bm{o}_t)$ (center), and \textit{freely generated} latent activity from the trained DSR model ${\bm{z}}_t=F_{\theta}(\bm{z}_{t-1})$ (bottom). 
  A linear operator $\bm{B}$ was optimized to align ground truth states $\bm{X}$ of the Lorenz-96 (not seen during training!) and states $\tilde{\bm{Z}}$ 
  encoded from the ordinal data via linear regression ($\bm{B} = \left( \tilde{\bm{Z}}^T \tilde{\bm{Z}} \right)^{-1} \tilde{\bm{Z}}^T \bm{X}$), and used to project the freely generated activity of the shPLRNN into the observation space of the Lorenz-96 system. \textbf{c}, left: Example of ground truth (orange) and freely generated (blue) activity, bearing the same temporal structure. Right: Aligned ground truth ($\bm{x}_t$) and encoded latent states ($\tilde{\bm{z}}_t$) as in b for one example unit. Note that the ground truth states $\bm{x}_t$ have \textit{never been seen by the model during training} but are only 
  inferred from the ordinal observations via the MTF encoder $p(\tilde{\bm{z}}_t|\bm{o}_t)$, yet still overlap almost perfectly.} \label{fig:lorenz96}
\end{figure}

\end{document}